\pdfoutput=1

\documentclass[11pt]{article}

\usepackage[final]{acl}

\usepackage{times}
\usepackage{latexsym}

\usepackage[T1]{fontenc}

\usepackage[utf8]{inputenc}

\usepackage{microtype}

\usepackage{inconsolata}

\usepackage{graphicx}

\usepackage{algorithm}
\usepackage[noend]{algpseudocode}

\usepackage{xcolor}
\usepackage{amsfonts}       
\usepackage[]{mdframed}
\usepackage{alltt}
\usepackage{fancyvrb}
\usepackage{fvextra}
\usepackage{amsmath} 
\usepackage{xspace}
\usepackage{subfig}  
\usepackage{multirow}
\usepackage{booktabs}

\usepackage{siunitx}
\usepackage{balance}
\definecolor{correct}{RGB}{150, 200, 50}

\newcommand{\standard}{$R^{\text{st}}$\xspace}
\newcommand{\standardRR}{$R^{\text{st}}_{+}$\xspace}
\newcommand{\skewed}{$R^{\text{sk}}$\xspace}
\newcommand{\skewedRR}{$R^{\text{sk}}_{+}$\xspace}
\newcommand{\accu}{$acc_{u}$\xspace}
\newcommand{\accg}{$acc_{g}$\xspace}
\newcommand{\acct}{$acc$\xspace}

\title{The Distracting Effect: Understanding Irrelevant Passages in RAG}

\author{
 \textbf{Chen Amiraz\textsuperscript{1}\thanks{\texttt{chen.amiraz@tii.ae}}},
 \textbf{Florin Cuconasu\textsuperscript{1,2}}\thanks{Work conducted while FC being a research intern at TII.}\thanks{\texttt{cuconasu@diag.uniroma1.it}},
 \textbf{Simone Filice\textsuperscript{1}\thanks{\texttt{filice.simone@gmail.com}}},
 \textbf{Zohar Karnin\textsuperscript{1}\thanks{\texttt{zohar.karnin@tii.ae}}}
\\
\\
 \textsuperscript{1}Technology Innovation Institute,
 \textsuperscript{2}Sapienza University of Rome
}

\begin{document}
\maketitle
\begin{abstract}
A well-known issue with Retrieval Augmented Generation (RAG) is that retrieved passages that are irrelevant to the query sometimes distract the answer-generating LLM, causing it to provide an incorrect response. In this paper, we shed light on this core issue and formulate the {\em distracting effect} of a passage w.r.t. a query (and an LLM). We provide a quantifiable measure of the distracting effect of a passage and demonstrate its robustness across LLMs. 
Our research introduces novel methods for identifying and using \textit{hard} distracting passages to improve RAG systems. By fine-tuning LLMs with these carefully selected distracting passages, we achieve up to a 7.5\% increase in answering accuracy compared to counterparts fine-tuned on conventional RAG datasets. Our contribution is two-fold: first, we move beyond the simple binary classification of irrelevant passages as either \textit{completely unrelated} vs. \textit{distracting}, and second, we develop and analyze multiple methods for finding hard distracting passages. To our knowledge, no other research has provided such a comprehensive framework for identifying and utilizing hard distracting passages.
\end{abstract}

\section{Introduction}
Retrieval Augmented Generation (RAG) is a key method to enable Large Language Models (LLMs) to solve knowledge-intensive tasks such as question-answering \cite{chen2017reading, petroni2021kilt}. Adding retrieved passages to the prompt of an LLM is shown to ground the LLM response, significantly reducing hallucinations \cite{fan2024survey}. 

Despite its advantages, retrieved content can sometimes lead to problematic behavior. Retrieval is not always successful, and in many such cases, the prompt includes distracting passages \cite{li2023large, yoran2024making}. As described by \citet{cuconasu2024power}, distracting passages contain irrelevant yet semantically related information that may mislead the LLM and thus hurt answer generation. 
Various solutions were proposed to handle such problematic retrieval results: based on Chain-of-Thought \cite{yu2023chain, weiinstructrag2024}, via LLM fine-tuning  \cite{yoran2024making, jin2024long}, and via dedicated inference procedures \cite{asaiself}.

This line of research raises a key question: \emph{how to evaluate the distracting effect of a passage on an LLM with respect to a query?} 

We begin tackling this question by defining a quantifiable measure of a passage’s distracting effect with respect to a query and an LLM. Our definition isolates the effect of the passage itself, which enables to decouple the influence of other passages. The distracting effect is inherently LLM-dependent, as different models may be affected by different passages. Despite this potential difference, we show that the distracting effect property is in fact quite robust to the LLM choice in that the scores have high correlations across LLMs. 
We further validate the robustness of our measure by showing that it translates to downstream RAG quality, specifically by demonstrating that the higher a passage’s distracting effect, the more it reduces accuracy when included in the prompt alongside the gold passage.

%
%

With the measure in hand, we move to study the distracting effect of retrieved passages. We corroborate results from previous studies \cite{jin2024long} showing that the irrelevant results obtained from stronger retrievers are more distracting when compared to weak retrievers. This phenomenon provides additional motivation to our study since retrievers will grow stronger with time, resulting in passages with a larger distracting effect. Our analysis reveals an additional, related observation: higher-ranking irrelevant results are more likely to be distracting.


To allow for test sets that reflect unseen data and/or training sets allowing for generalizable models, we aim to obtain distracting passages to all queries, including those where standard retrieval fails. Such failures occur either when the retriever does not return distracting passages or when no such passages exist in the corpus (e.g., in niche topics or small corpora). To address the first case, we define a skewed retriever tuned to provide passages related to the query but unrelated to its answer. For the second, we define several categories of distracting passages, inspired by \citet{basmov2024llms, abdumalikov2024answerability}, and generate passages for each category using a strong LLM.

%
%

We demonstrate the effectiveness of this diverse collection of methods by analyzing their ability to jointly provide highly distracting passages to queries in public question-answering benchmarks. We show that for a non-negligible fraction of queries, the joint collection of methods allows for much more distracting passages compared to any single method, in particular that of standard retrieval.


We finish by demonstrating the usefulness of our techniques for collecting distracting passages; using these passages we build a training set used to fine-tune an LLM on a question-answering task. We observe that a fine-tuned LLM based on our training set achieves superior results to one fine-tuned on an analogous training set obtained via standard retrieval. 

Concluding, our contributions are as follows: (1) We formalize a core issue in RAG, that of distracting passages, providing a formal definition and evaluation method for such passages (2) We present diverse techniques to obtain such distracting passages, (3) We demonstrate the value of distracting passages by building an effective RAG training set.

\section{Related work}
\paragraph{Analysis of Irrelevant Passages.} 
One line of research in RAG focuses on analyzing different types of irrelevant passages. A passage is considered relevant if it contains the correct answer (or part of it) and provides useful context for answering the query. \citet{cuconasu2024power} classify irrelevant passages as either random or distracting, showing that while random passages do not degrade answer quality, distracting passages do. We adopt the term distracting but extend it beyond classification by treating distraction as a continuous property and providing concrete methods to quantify it.
\citet{basmov2024llms} analyze the LLM's ability to answer questions when the provided reference passages contain hypothetical statements or statements that contradict its parametric knowledge. They show that in both cases performance can significantly drop. We make use of this categorization (among others) of distracting passages in order to synthetically generate diverse types of distracting passages. \citet{jin2024long} show that irrelevant passages returned by strong retrievers are more distracting than those obtained by weak retrievers by showing that RAG systems tend to make more mistakes when given the former rather than the latter. In our analysis, we provide additional ways to measure how distracting a passage is, and corroborate this conclusion.

\paragraph{Obtaining Distracting Passages.} 
To our knowledge, in the context of answer generation, the existing solutions to obtain distracting passages are all based on retrieval. 
The dominant technique is by obtaining top-ranked passages that are not the ground-truth passage, e.g., \cite{yoran2024making}\footnote{Their precise method is in fact not the top passages, but rather a uniform random set of $k$ passages out of the top $K>k$.}. \citet{abdumalikov2024answerability} went beyond standard retrieval by generating synthetic passages either containing the question but not the answer, or the answer and not the question. They do so in an effort to teach an LLM to abstain when needed.

In the context of retrieval and reranking, there is a rich line of work exploring methods for obtaining hard negatives, i.e., passages that are irrelevant to the query but seem relevant to the retrieval or reranking system. For the retrieval problem, the dominating method is that of contrastive learning \cite{xiongapproximate2021}, in which the hard negatives are implicitly found by the training method, but this technique is only possible when the pairwise similarity is a simple function (e.g., inner product) and is inapplicable for cross-encoders typically used for reranking. Here, the methods are variations of taking the top results from an existing retriever or reranker that are not the ground truth passage. Other than the difference in their definition (hard negatives are defined w.r.t. the ranker, not the answer generator), another key distinction is how to deal with false negatives, meaning passages that are not labeled as relevant but are in fact relevant. \citet{moreira2024nv} discuss such methods that discard negative candidates whose score is larger than some threshold, either fixed or based on the score of a known positive example. In our setting, this is less of a problem given that we have a ground-truth answer. This additional information allows a more accurate filtering of false negatives. Due to this, we focus on additional methods of providing candidate distracting examples rather than ways to filter false negatives. Another notable recent work in the area of retrieval is by \citet{weller2024promptriever} that train a promptable retriever that can retrieve passages relevant to a query and an instruction. Here, the authors synthesize hard negative passages that match the query but do not match the instruction. Our setting is fundamentally different in the definition of a negative example, and due to this we use completely different methods to generate such examples.

\paragraph{Robust Answer Generation.} Closely related to the above, another relevant area is that of building answer generation methods that can handle irrelevant and distracting passages. One approach is to have a chain-of-thought process, either via a prompt or fine-tuning in which the LLM identifies the relevant passages \cite{yu2023chain, yan2024corrective, luo2023sail, weiinstructrag2024}. A similar approach is taken by \citet{asaiself}. They provide an entire RAG system, but one of its components indirectly decides whether a passage is relevant by generating an answer with it and measuring its faithfulness to the passage. Another approach is to fine-tune LLMs to answer questions when coupled with both relevant and irrelevant passages. \citet{linra2024} and \citet{jin2024long} do this with passages obtained from a standard retrieval system. \citet{yoran2024making} do the same, but add examples where the passages are intentionally irrelevant, specifically they are sampled from the top results rather than taking the top results. 

\section{Distracting Passages} \label{sec:method}
An informal definition of the distracting effect of a passage w.r.t a query and LLM is: given a query $q$ and a passage $p$ that is irrelevant to $q$, how likely is an LLM to be distracted by the passage? In this section, we provide a concrete measure for the distracting effect of a passage, then move to describe different methods to obtain distracting passages. 


\subsection{Measuring the Distracting Effect}
\label{sec:score_candidates}

For the formal test, we build a prompt from $q$ and $p$ where we ask the LLM to answer $q$ based on the passage and abstain (output ``NO-RESPONSE'') if the passage does not contain an answer to $q$. The precise prompt is given in Figure~\ref{fig:prompt dist test} and all the implementation details are described in Appendix \ref{sec:appendix_details_distracting}. We compute distracting effect $\text{DE}_q(p)$ of an irrelevant passage $p$ for query $q$ as the probability of the LLM not abstaining:
\begin{equation}
\label{eq:distracting_effect}
    \text{DE}_q(p) = 1 - p^{\text{LLM}}(\text{NO-RESPONSE}|q,p)
\end{equation}

Alternatives to this test could be comparing the answer of the LLM with vs. without the passage, or building a prompt that also includes a relevant passage and checking if $p$ changes the response. While these approaches are viable, our DE$_q(p)$ score offers several key advantages: 
(1) as a probability measure bounded between 0 and 1, it provides an easily interpretable metric of distraction, 
(2) since this score leverages the LLM's intrinsic ability to recognize relevant information, it can be applied beyond question-answering to any task where distinguishing between relevant and distracting information is crucial, 
(3) it applies to the LLM being tested, without relying on an expensive reference model, 
(4) it does not require additional passages nor assumptions about the LLM's parametric memory, and (5) it has a relatively cheap implementation cost, as it simply requires the LLM to process the prompt without generating any new tokens. 

In our analysis, we interpret DE$_q(p)$ as a relative ranking score for passages associated with the same query. This comparative usage helps mitigate concerns that an LLM might ignore the instruction and rely instead on its parametric knowledge to answer the question. Even if such behavior might occur to some extent, it would affect all passages similarly for a given query, preserving the validity of DE$_q(p)$ as a measure of their relative distracting effects.



\subsection{Obtaining Distracting Passages}\label{sec:getting_distractors}
Here, we have two approaches. The first is to retrieve candidates, and the second is to generate them using an LLM. The former method will provide examples closer to those observed at inference time. However, the synthetic examples have the potential to add robustness to the system for rarely observed types of distracting passages. Additionally, for small corpora, distracting examples may be impossible to achieve for many queries, e.g., when they discuss a topic present in a single document. Here, the synthetic examples are key for a robust system.

A key challenge in learning-to-rank settings is distinguishing truly irrelevant passages from false negatives, i.e., passages that are mistakenly treated as irrelevant but actually contain useful information. This issue is often addressed by excluding top-ranked candidates or those with high relevance scores. In our case, however, we aim to ensure that the passages we obtain (by all methods) are indeed irrelevant, which is more straightforward thanks to access to the ground truth answer. Specifically, we use the NLI model of \citet{honovich2022true} in the following manner. A passage is considered relevant to a query if it explicitly includes the ground truth answer or entails the hypothesis ``the answer to \{question\} is \{answer\}'' given the passage as the premise. We exclude such passages when computing distracting effect scores.

\subsubsection{Retrieving Distracting Passages}
\label{sec:retrieval}
The idea here follows the intuition that irrelevant passages ranked in a top position by a retrieval system are likely to have a large distracting effect. For different retrieval systems, we obtain the resulting passages and exclude the relevant ones to obtain either a single (the remaining top-ranked) or a ranked list of candidate passages.


In addition to standard retrieval, we consider a modified version of (dense) retrieval that we call answer-skewed retrieval. A dense retriever is defined via embedding functions $E_Q, E_D$ mapping a query/document into an embedding space. While keeping the document embedding the same, we modify the query embedding as follows: for a query $q$ coupled with a ground-truth answer $a$ we define
\begin{equation}
\label{eq:sub}
    E^{\text{sub}}(q,a) = E_Q(q) - \lambda E_D(a)
\end{equation}
and 
\begin{equation}
\label{eq:proj}
    E^{\text{proj}}(q,a) = E_Q(q) - \lambda \frac{\left\langle E_Q(q), E_D(a) \right\rangle E_D(a)}{\left\|E_D(a)\right\|^2}
\end{equation}
The former subtracts the answer embedding from the original query embedding, and the latter projects it. 
These formulas are the arithmetic way to express the idea of retrieving a document that is related to the query but unrelated to the answer. The hyper-parameter $\lambda$ determines how aggressively we wish to exclude documents related to the answer.

\subsubsection{Generating Distracting Passages}
\label{sec:generation}
Here, we use a categorization of different types of distracting passages inspired by \citet{basmov2024llms, abdumalikov2024answerability}. For each type, we employ few-shot learning, i.e., build a prompt containing a handful of query and distracting passage pairs (see Appendix~\ref{sec:distracting prompts}). We then use a strong LLM 
to generate a passage of the corresponding distracting category. The categories used are:
    \paragraph{Related Topic} A passage discussing a topic highly related to the question, but that does not contain the answer. E.g., for ``When was Abraham Lincoln born?'', ``Robert Todd Lincoln, the eldest son of President Abraham Lincoln, was born August 1, 1843.''.
    The generator of this type of distracting documents is referred to as $G^{\text{rel}}$.
    
    \paragraph{Hypothetical} A passage discussing the question in a hypothetical situation in which the answer is different. E.g., for ``What is a traditional gift for a 5th anniversary?'' ``In ancient Roman times, couples would go on a week long hunting trip on their 5th anniversary''.
    The generator of this type of distracting documents is referred to as $G^{\text{hypo}}$.
    
    \paragraph{Negation} A passage providing a wrong answer, but in negation. E.g., ``It is a common misconception that students do not pay tax on earnings''.
    The generator of this type of distracting documents is referred to as $G^{\text{neg}}$.
    
    \paragraph{Modal Statement} A passage providing a wrong answer following a disclaimer that the statement is not certain. E.g., ``The Pyramids may have been built via employing a sloping and encircling embankment of brick, earth, and sand.''.
    The generator of this type of distracting documents is referred to as $G^{\text{modal}}$.


\section{Analyzing the Distracting Effect}
In this section, we analyze the different techniques for obtaining distracting candidates discussed in Section~\ref{sec:method}, and we show the benefit of jointly using different methodologies to create sets of highly distracting passages. Furthermore, we show how highly distracting passages can affect the LLM response quality even when a relevant document is present in the prompt.

\begin{figure*}[t!]
    \centering
    \includegraphics[width=\linewidth]{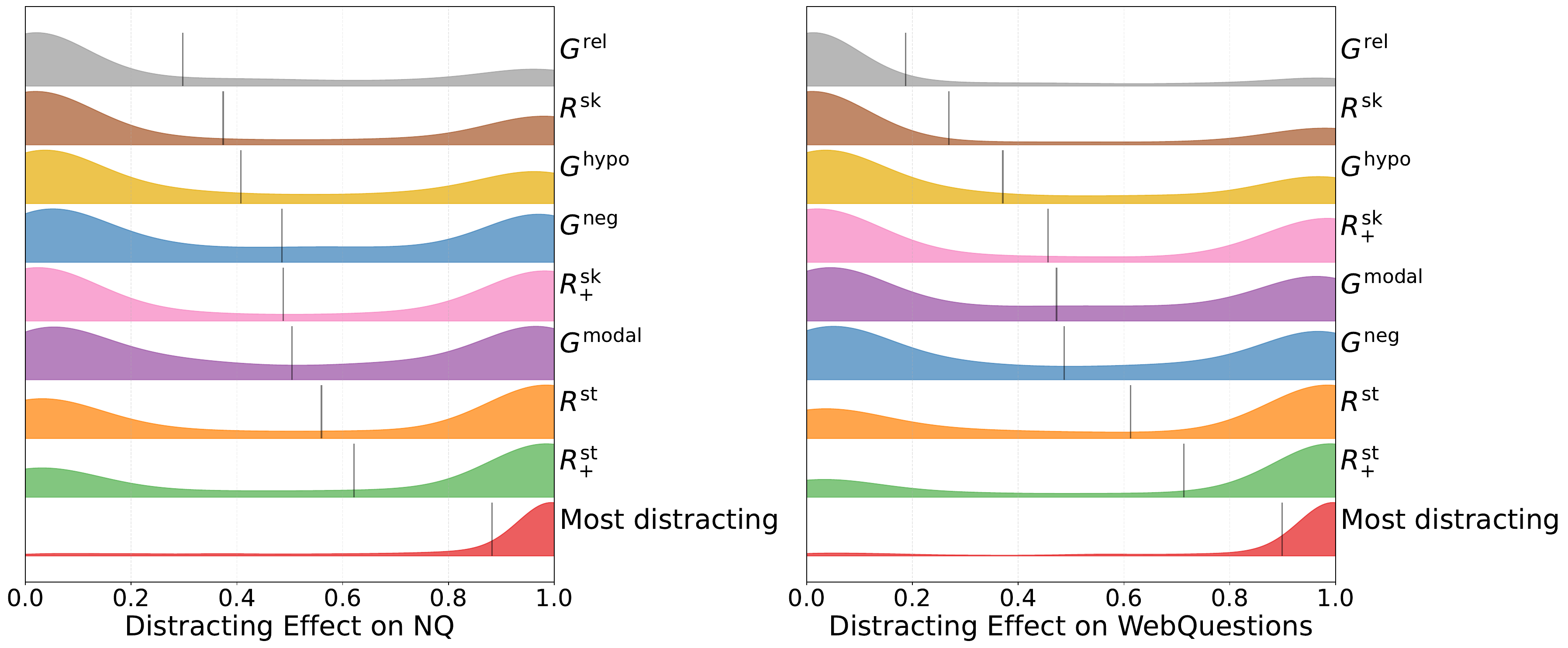}
    \caption{Distribution of distracting effect for passages obtained through different methods, as measured by Llama-3.1-8B. Methods are ordered by their mean distracting effect (shown by vertical black lines), with higher means indicating a greater ability to distract the model.}
    \label{fig:distr_prob_distr}
\end{figure*}

\subsection{Experimental Setting} \label{sec:effect exp setting}
\paragraph{Benchmarks.} We make use of the following commonly used public question-answering benchmarks: NQ \cite{kwiatkowski2019natural}, PopQA \cite{mallen2023not}, TriviaQA \cite{joshi2017triviaqa} and WebQA \cite{berant-etal-2013-semantic}. We took a sample of 2000 queries from NQ and 1000 from the rest. We filtered out instances without relevant passages among the retrieved ones, to enable tests related to ground-truth passages. This resulted in 1926, 950, 987, and 837 queries for the respective datasets. These benchmarks all come with a reference answer, which we use to assess the correctness of the generated answers. In particular, we adopt a common variant of the Exact Match metric, where we classify a generated answer as correct if it includes the ground-truth answer as a substring (e.g., if the answer is ``Washington'' and the generated answer is ``George Washington'', it is considered correct). This procedure will be generally referred to as answer accuracy in our experiments.


\paragraph{Compared Methodologies.} We compare the distracting effects of the passages obtained by using the methodologies discussed in Section \ref{sec:method}. For the retrieval-based methods, we index the Wikipedia dump of 20 December 2018 \cite{gao2023retrieval} using Pinecone vector-DB\footnote{\url{https://www.pinecone.io/}} and the E5-base embedding model \cite{wang2022text} with 768 embedding dimension. We also explored the answer-skewed retriever\footnote{Among tested configurations, formulation \ref{eq:sub} with $\lambda=1$ performed best. See Section \ref{sec:appendix_skewed_retr} for further details.} from Section \ref{sec:retrieval}. 
Hereafter, we use \standard to refer to a standard dense retriever and \skewed to refer to its answer-skewed counterpart. We evaluate both solutions with and without re-ranking their top-20 passages using the cross-encoder BAAI BGE-M3-v2\footnote{\url{https://huggingface.co/BAAI/bge-reranker-v2-m3}} \cite{chen2024bge}. We use \standardRR and \skewedRR to refer to the retrieval models followed by the re-ranking module.  
Regarding the four generation methods discussed in Section \ref{sec:generation}, we use Claude 3.5 Sonnet V2.0 via AWS Bedrock as the backbone LLM.

\paragraph{Evaluated LLMs.} To assess the distracting effect of a candidate passage we use formula \ref{eq:distracting_effect} with various LLMs. We consider open-sourced LLMs ranging from 3B to 70B parameters, specifically the \textbf{instruct-based} version of Llama-3.2-3B, Llama-3.1-8B, Llama-3.3-70B \cite{grattafiori2024llama3herdmodels}, Falcon-3-3B, Falcon-3-7B \cite{Falcon3}, Qwen-2.5-3B, and Qwen-2.5-7B \cite{qwen2025qwen25technicalreport}. 


\subsection{Distracting Effect of Retrieved Passages}
\label{sec:distraction_of_retrieval_methods}
\begin{figure}[t]
    \centering
    \includegraphics[width=\linewidth]{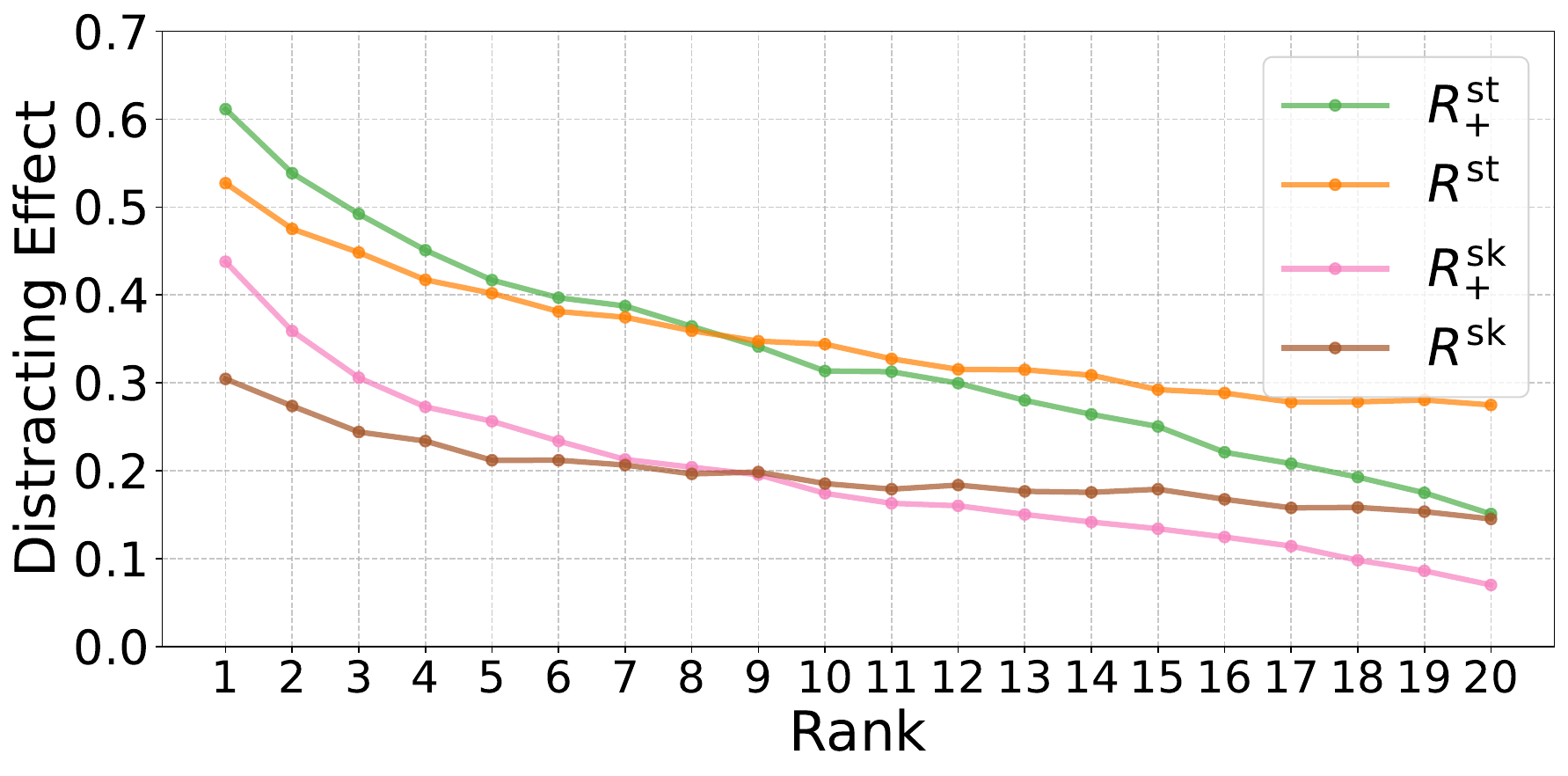}
    \caption{Average distracting effect at different rank positions for various retrieval methods. Results are shown for Llama-3.1-8B, averaged across datasets. Higher-ranked passages consistently demonstrate greater potential to mislead the model. Similar trends were observed across all tested LLMs (see Figures~\ref{fig:all_distracting_rank} and \ref{fig:llama70b_distracting_rank}).}
    \label{fig:distr_by_rank}
\end{figure}

In this experiment, we compute the average distracting effect of the irrelevant passages retrieved using the various retrieval methods. Figure \ref{fig:distr_by_rank} shows how the distracting effect (averaged across the four datasets we consider) varies at different ranking positions. All methods exhibit the same decreasing trend. 
A notable conclusion is that standard retrieval pipelines, while attempting to bring relevant passages to top positions, tend to favor passages with high distracting effects over passages with low distracting effects. 

Another important observation arises when analyzing the effect of the reranking on the top positions (e.g., top 5). In these positions, using the reranking module consistently increases the average distracting effect. We argue that the irrelevant passages that are retrieved are the ones that \textit{fool} the retrieval pipeline, and that also have the potential of distracting the LLM. While adding a reranking module enhances the capabilities of the retrieval pipeline, this actually amplifies the problem – passages that successfully pass through this additional reranking stage are even more likely to mislead the LLM during response generation.

\subsection{Comparing Distracting Effects of Different Methodologies}
\label{sec:compare_distracting_effects}

In these experiments, we compare all the methods discussed in Section \ref{sec:method}. For retrieval-based methods, we consider only the first non-relevant passage for each query, which is expected to be the most distracting retrieved one according to Figure \ref{fig:distr_by_rank}. Similarly, for a fair comparison, for each query, we use each of the four methods discussed in Section \ref{sec:generation} to generate a single distracting passage.

Figure \ref{fig:distr_prob_distr} reports the probability distribution of the distracting effect over Llama-3.1-8B of the passages obtained by using the different methods, computed on NQ and WebQuestions. Results on TriviaQA and PopQA follow the same trend and are described in Appendix \ref{sec:appendix_distracting_others}. 
The probability distributions are all skewed towards extreme values showing the LLM tendency to always respond with high confidence, even when wrong. The probability distributions of the other LLMs, described in Appendix \ref{sec:appendix_distracting_others}, follow similar trends. 
An important aspect to notice is that 
the relative distracting effect provided by the different methods is quite stable: \standardRR and \standard are among the top-distracting approaches in all datasets, and similarly, $G^{\text{rel}}$ and \skewed are among the poorest-performing methods across all datasets. This suggests that the inherent strengths and weaknesses of these methodologies transcend the specific characteristics of individual datasets.

Regarding the retrieval-based methods, the results are in line with Section \ref{sec:distraction_of_retrieval_methods}, with \standard consistently providing passages having a higher distracting effect than \skewed; in both cases, reranking leads to more distracting passages. 

%
%
Among the generation-based methods, $G^{\text{modal}}$ appears on average the most promising to produce distracting passages; on the opposite, the passages generated by $G^{\text{rel}}$ are the least distracting ones. 

\begin{figure}[t]
    \centering
    \includegraphics[width=\linewidth]{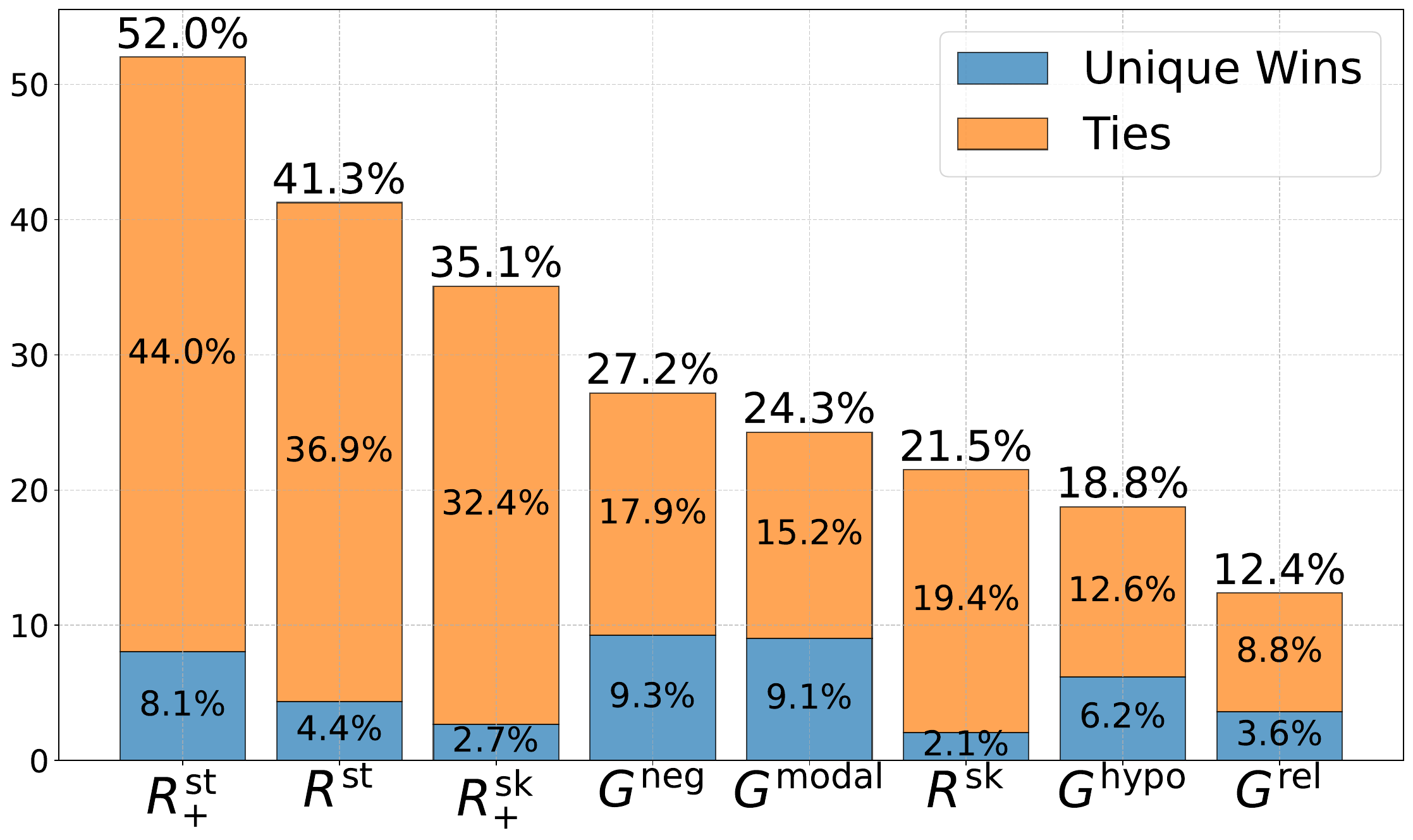}
    \caption{Percentage of queries where each method provides the most distracting passage for Llama-3.1-8B. In blue are the times when no other method reaches the same distracting effect, in orange the percentage of times the highest score is shared with other methods. Similar trends were observed across all tested LLMs (see Figures~\ref{fig:all_best_distractors} and \ref{fig:llama70b_best_distractors}).}
    \label{fig:best_distractors}
\end{figure}

Finally, the last probability distribution in each subfigure of Figure \ref{fig:distr_prob_distr} refers to the case where for each query we systematically select the most distracting passage among the ones obtained with the various methods. In this case, the probability is mostly distributed on high distracting effect values, demonstrating that the joint usage of different methodologies leads to significantly more distracting passages than the ones obtained by any of the individual methods. Figure \ref{fig:best_distractors} allows to better understand the contribution of each method. The vertical bars represent the percentage of queries (from the four datasets we analyze) where each method provides the most distracting passage. In blue are the times when no other method reaches the same distracting effect, in orange the percentage of times the highest score is shared with other methods (a difference below 0.01 is considered a tie). 

In line with the probability distributions observed in Figure \ref{fig:distr_prob_distr}, \standardRR is the method providing the most distracting passage for the highest number of queries. Nevertheless, for $\sim$48\% of the queries, other methods produce more distracting passages. Overall, all methods provide their unique contribution, which is particularly remarkable not only for \standardRR but also for $G^{\text{modal}}$, $G^{\text{neg}}$, and $G^{\text{hypo}}$, demonstrating that combining retrieval and generation-based solutions is beneficial to obtain highly distracting passages for a set of queries. 

\begin{figure}[t]
    \centering
    \includegraphics[width=\linewidth]{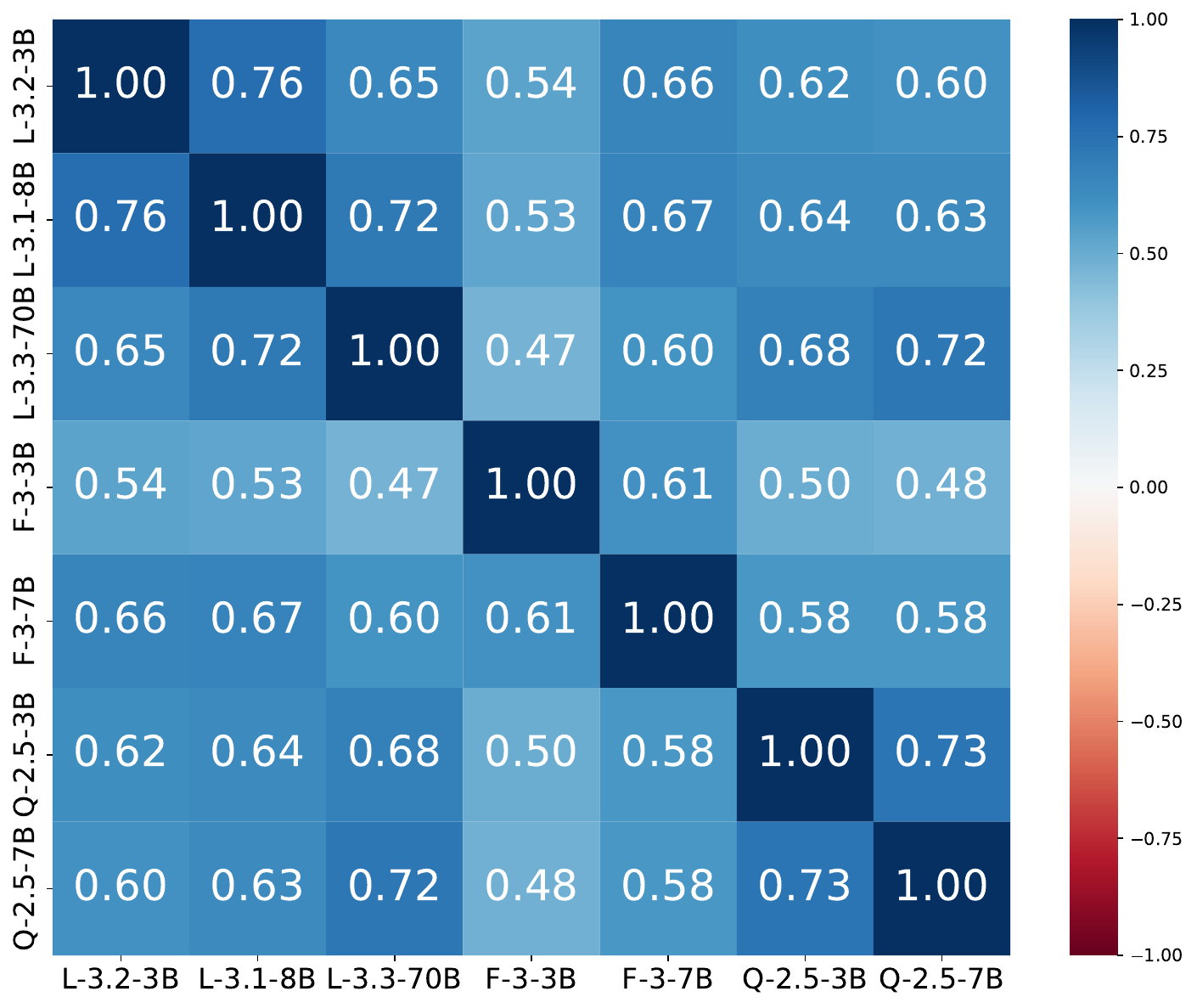}
    \caption{Spearman correlation of distracting effect computed using different LLMs (abbreviated, e.g., Llama $\rightarrow$ L). The strong correlations suggest that the distracting effect of a passage is relatively consistent across models despite architectural differences.}
    \label{fig:spearman}
\end{figure}

The experiments reported so far study the distracting effect on Llama-3.1-8B, however, we observe very similar trends with other LLMs (details in Appendix \ref{sec:appendix_distracting_others}): some LLMs are more distractable than others, but overall the relative effectiveness of the various methods is very similar. Figure \ref{fig:spearman} provides deeper insights into how the distracting effect depends on the LLM used to compute it. We observe very high Spearman correlation scores between the distracting effects computed using different LLMs; this means that the LLMs we analyze share the same weaknesses and tend to be more distracted by the same set of passages. We argue that the distracting effect of a passage is an intrinsic characteristic of the passage itself and that it does not depend much on the LLM.

\begin{table}[t]
\centering
\resizebox{\columnwidth}{!}{
\begin{tabular}{@{}lccc@{}}
\toprule
\multicolumn{1}{c}{\textbf{LLM}} & \textbf{Only Gold} & \textbf{Gold + WD}                     & \textbf{Gold + HD} \\ \midrule
Llama-3.2-3B            & 82.6      & 79.4                          & 71.5      \\
Llama-3.1-8B            & 80.6      & \underline{80.1} & 73.9      \\
Llama-3.3-70B            & 81.1     & 80.1 & 75.2      \\
Falcon-3-3B             & 78.5      & 74.1                          & 67.1      \\
Falcon-3-7B             & 84.1      & 81.5                          & 73.3      \\
Qwen-2.5-3B             & 80.9      & 75.5                          & 69.4      \\
Qwen-2.5-7B             & 82.4      & 80.4                          & 73.7      \\ \bottomrule
\end{tabular}
}
\caption{Answer accuracy when prompting the LLM with the gold passage only, gold passage with a weak distracting passage (WD), and gold passage with a hard distracting passage (HD). Values that are NOT underlined are different in a statistically significant way w.r.t. the gold-only case (Wilcoxon test with p-value < 0.01).}
    \label{tab:gold_with_distracting}
\end{table}

\subsection{Prompting with Relevant and Distracting Passages}
\label{sec:acc_drop}
So far, we measured the distracting effect of an irrelevant passage as the probability of the LLM not abstaining when prompted with it. A question we should still answer is: how do distracting passages affect the LLM accuracy when added to a prompt already containing a relevant passage?
To answer this question we categorize irrelevant passages as \emph{hard} distracting if associated with a distracting effect higher than 0.8. Similarly, we consider \emph{weak} distracting those passages having a distracting effect lower than 0.2. 
An example of hard and weak distracting passages is shown in Figure~\ref{fig:prompt1_hard_vs_weak}, with additional examples obtained through different methods in the Appendix (Figures~\ref{fig:prompt2_hard_vs_weak}, \ref{fig:prompt3_hard_vs_weak}, and \ref{fig:prompt4_hard_vs_weak}).

As evident from the probability distributions in Figure \ref{fig:distr_prob_distr}, these two intervals account for most of the probability mass (e.g., 72\% for Llama-3.1-8B). Table \ref{tab:gold_with_distracting} shows how accuracy on the NQ dataset drops when the relevant passage (i.e., we use the gold passage available in the NQ dataset) is combined with a distracting one in the LLM prompt. 
Since LLMs are known to be affected by positional bias \cite{liu2023lostmiddlelanguagemodels}, we compute both orders (i.e., gold followed by distracting, and distracting followed by gold) and report the average accuracy. Both weak and hard distracting passages affect accuracy when compared to having only a gold document in the prompt, however, the impact of hard distracting passages is significantly larger, making the accuracy drop from 6 to 11 accuracy points, depending on the LLM. Notably, this degradation persists even for larger LLMs, like Llama-3.3-70B. This analysis confirms the reliability of the strategy we adopt for scoring distracting passages.

\begin{figure*}[t!]
\begin{mdframed}[font=\footnotesize]
\begin{Verbatim}[breaklines=true, breaksymbol=, commandchars=\\\{\}]
\textbf{Question}: What movie has the song on the road again?

\textbf{Gold Answer}: \textcolor{correct}{Honeysuckle Rose}
\end{Verbatim}
\noindent\rule{\textwidth}{1pt}
\textbf{Relevant Passage}
\begin{Verbatim}[breaklines=true, breaksymbol=, commandchars=\\\{\}]
(Title: On the Road Again (Willie Nelson song)) 
The song , about life on tour , came about when the executive producer of the film \textcolor{correct}{Honeysuckle Rose} approached Nelson about writing the song for the film 's soundtrack . '' On the Road Again '' became Nelson 's 9th Country & Western No. 1 hit overall ( 6th as a solo recording act ) in November 1980 , and became one of Nelson 's most recognizable tunes . In addition , the song reached No. 20 on the Billboard Hot 100 , and No. 7 on the Adult Contemporary chart . It was his biggest pop hit to that time and won him a Grammy Award for Best Country Song a year later .
\end{Verbatim}
\noindent\rule{\textwidth}{1pt}
\begin{minipage}[t]{0.48\textwidth}
\textbf{Hard Distracting} $G^{\text{modal}}$ 
\newline \newline
Distracting Effect: \newline
\begin{tabular}{@{}l@{\hspace{0.2em}}l@{}}
{\tt - Llama-3.2-3B:} & {\tt 0.9680} \\
{\tt - Llama-3.1-8B:} & {\tt 0.3125} \\
{\tt - Llama-3.3-70B:} & {\tt 0.0430} \\
{\tt - Falcon-3-3B:} & {\tt 1.0000} \\
{\tt - Falcon-3-7B:} & {\tt 0.3789} \\
{\tt - Qwen-2.5-3B:} & {\tt 1.0000} \\
{\tt - Qwen-2.5-7B:} & {\tt 0.4375}
\end{tabular}
\begin{Verbatim}[breaklines=true, breaksymbol=, commandchars=\\\{\}]
(Title: Classic Songs in Films) 
Many people believe, though it's not actually correct, that Willie Nelson's iconic song 'On The Road Again' first appeared in the 1980 film '\textcolor{red}{Smokey and the Bandit II}.' Some music historians have suggested that this misconception arose because the film's themes of truck driving and life on the road seemed to perfectly match the song's message. The song's road-trip vibe made it a natural fit for many movies, but this particular connection is just a popular misconception.

\textbf{Generated Answer}: \textcolor{red}{Smokey and the Bandit II}
\end{Verbatim}
\end{minipage}
\hfill
\vrule width 1pt
\hfill
\begin{minipage}[t]{0.48\textwidth}
\textbf{Weak Distracting} $G^{\text{rel}}$ 
\newline \newline
Distracting Effect: \newline
\begin{tabular}{@{}l@{\hspace{0.2em}}l@{}}
{\tt - Llama-3.2-3B:} & {\tt 0.0234} \\
{\tt - Llama-3.1-8B:} & {\tt 0.0312} \\
{\tt - Llama-3.3-70B:} & {\tt 0.0000} \\
{\tt - Falcon-3-3B:} & {\tt 0.3371} \\
{\tt - Falcon-3-7B:} & {\tt 0.0117} \\
{\tt - Qwen-2.5-3B:} & {\tt 0.0000} \\
{\tt - Qwen-2.5-7B:} & {\tt 0.0000}
\end{tabular}
\begin{Verbatim}[breaklines=true, breaksymbol=, commandchars=\\\{\}]
(Title: Willie Nelson Hits) 
Country music legend Willie Nelson has recorded numerous hits throughout his career that have been featured in various films. His iconic song 'Always on My Mind' was prominently featured in the 1982 film 'The Soldier' and became one of his signature performances. Another of his classics, 'Blue Eyes Crying in the Rain,' was used effectively in the 2005 film 'Brokeback Mountain' during a pivotal scene.


\textbf{Generated Answer}: \textcolor{correct}{Honeysuckle Rose}
\end{Verbatim}
\end{minipage}
\end{mdframed}
\vspace{-0.4cm}
\caption{Example showing Falcon-3-3B's responses in two scenarios using the prompt template in Figure \ref{fig:prompt_ft}: relevant passage + hard distracting and relevant passage + weak distracting. \textbf{Left}: When the relevant passage is followed by a hard distracting passage (generated by $G^{\text{modal}}$ and classified as hard for 3B models due to their distracting effect $>0.8$), Falcon-3-3B answers ``Smokey and the Bandit II'' instead of ``Honeysuckle Rose'', despite having access to the relevant information. \textbf{Right}: When the relevant passage is followed by a weak distracting passage (generated by $G^{\text{rel}}$), the model correctly answers ``Honeysuckle Rose''.}
\label{fig:prompt1_hard_vs_weak}
\end{figure*}

\section{Application: RAG Fine-Tuning} \label{sec:app}
We now move to make use of the distracting passages to train a robust generation component for RAG. For each query, we obtain through the methods described above, a set of highly distracting passages, as well as a relevant passage. We use these to build a prompt containing the query and passages, resulting in a training set for RAG. 

\subsection{Experimental Setting}\label{sec:ft_exp}
We adopt the same benchmarks as in Section~\ref{sec:effect exp setting}. We use 800 queries from NQ to build training data. Each instance in our training data is a $(q,a^*,P)$ triplet consisting of a query, a ground-truth answer, and a list of 5 passages. We use three strategies to collect the passages in $P$: \textbf{Retrieve} – we use our retrieval pipeline (see Section \ref{sec:effect exp setting}) without re-ranker and fetch the top 5 ranked results; \textbf{Rerank} – same as retrieve, but in this case, we enable the re-ranking module; \textbf{Hard} – in 50\% of the cases we take the first relevant passage from the Rerank strategy and the most distracting 4 passages obtained by using the methods described in Section \ref{sec:getting_distractors}; in the remaining 50\% of the cases, we select the most distracting 5 passages obtained by using our methods. Finally, we shuffle the five passages to create $P$.

These strategies resulted in three corresponding training sets. We use the remaining queries from NQ to create an in-distribution (ID) test set. Additionally, we use the queries from PopQA, TriviaQA, and WebQA to create out-of-distribution (OOD) test sets.
Each of the resulting four test sets contains a balanced mixture of test cases from the \textit{Retrieve}, \textit{Rerank}, and \textit{Hard} strategies described above. 
We use the training sets to fine-tune\footnote{For further details, see Appendix \ref{sec:exp_setup}.} the instruct-based version of two LLMs, namely Llama-3.2-3B and Llama-3.1-8B, and compare results to their non-fine-tuned counterparts.
  




\subsection{Results}\label{sec:ft_results}

Table~\ref{tab:ft results} contains the test results for all 8 LLMs, corresponding to both the bases of Llama-3.2-3B and Llama-3.1-8B. In addition to overall accuracy, we report $acc_g$, $acc_u$ corresponding to accuracy over grounded examples, i.e., those that contain a relevant passage, and ungrounded examples that do not, respectively.

In all cases, training on \textit{Hard} examples results in major lifts over all baselines in all test sets: 5.3-16.1 absolute accuracy points for Llama-3.2-3B and 3.6-11.0 points for Llama-3.1-8B. 
We conjecture this is due to the added value of robustness to distracting passages in the case of ungrounded examples. Indeed, when the ground truth passage is in the prompt, a distracting passage can hurt, but to a limited effect (Table~\ref{tab:gold_with_distracting} shows drops from 6 to 11 accuracy). In contrast, when the answer is present only in the parametric memory of the LLM, a prompt with only distracting passages is much more likely to result in an error.
Due to space restrictions, we provide the test results on the different partitions of the test sets according to the passage collection method in Appendix \ref{sec:acc_table}. Results exhibit consistent behavior with major improvements on ungrounded examples across all slices.

For the overall accuracy, we see a clear advantage of {\em Hard} for Llama-3.2-3B across the board, especially for OOD datasets, with a lift of $6.7$ and $7.6$ $acc$ points for the TriviaQA and WebQA when compared to the baselines. For Llama-3.1-8B, since it is a stronger LLM with a lower margin of improvement, the results are closer to the baselines, though the overall performance is better for our technique in 3 out of 4 benchmarks. As before, the gain is much more significant for the ungrounded instances, but here it does come at a small expense of accuracy on grounded instances.

\begin{table}[t!]
    \centering
    \setlength{\tabcolsep}{4pt}
    \resizebox{\columnwidth}{!}{
    \begin{tabular}{@{}l l *{6}{S[table-format=2.1]}@{}}
    \toprule
    \multirow{2}{*}{\textbf{Test Set}} & \multirow{2}{*}{\textbf{Train Set}} & \multicolumn{3}{c}{\textbf{Llama-3.2-3B}} & \multicolumn{3}{c}{\textbf{Llama-3.1-8B}} \\
    \cmidrule(lr){3-5} \cmidrule(lr){6-8}     
     &  & \accu & \accg & \acct & \accu & \accg & \acct \\
    \midrule
    \multirow{4}{*}{NQ} 
        & \textit{None} & 15.2 & 51.4 & 37.9 & 12.8 & 56.7 & 40.3 \\
        & \textit{Retrieve} & 13.3 & \textbf{57.1} & 40.7 & 21.0 & 62.4 & 46.9 \\
        & \textit{Rerank} & 11.5 & 56.5 & 39.7 & 19.9 & \textbf{63.2} & 47.0 \\
        & \textit{Hard} & \textbf{21.4} & 55.6 & \textbf{42.8} & \textbf{32.0} & 59.8 & \textbf{49.4} \\ \midrule
    \multirow{4}{*}{PopQA} 
        & \textit{None} & 8.4 & 55.7 & 35.9 & 8.7 & 61.3 & 39.3 \\
        & \textit{Retrieve} & 8.8 & 62.6 & 40.1 & 16.6 & 73.2 & 49.5 \\
        & \textit{Rerank} & 9.6 & 60.3 & 39.1 & 18.0 & \textbf{75.1} & \textbf{51.2} \\
        & \textit{Hard} & \textbf{14.9} & \textbf{63.6} & \textbf{43.2} & \textbf{21.6} & 71.2 & 50.4 \\
    \midrule
    \multirow{4}{*}{TriviaQA} 
        & \textit{None} & 38.0 & 79.2 & 67.8 & 39.8 & 86.4 & 73.5 \\
        & \textit{Retrieve} & 36.9 & 79.4 & 67.6 & 56.8 & 87.1 & 78.7 \\
        & \textit{Rerank} & 33.3 & 76.7 & 64.7 & 58.4 & \textbf{87.5} & 79.4 \\
        & \textit{Hard} & \textbf{54.1} & \textbf{82.3} & \textbf{74.5} & \textbf{68.9} & 87.0 & \textbf{82.0} \\
    \midrule
    \multirow{4}{*}{WebQA} 
        & \textit{None} & 21.4 & 54.4 & 41.9 & 19.0 & 53.8 & 40.6 \\
        & \textit{Retrieve} & 20.7 & 55.1 & 42.1 & 28.4 & \textbf{59.9} & 48.0 \\
        & \textit{Rerank} & 20.3 & 54.1 & 41.3 & 30.4 & 59.6 & 48.6 \\
        & \textit{Hard} & \textbf{35.0} & \textbf{58.7} & \textbf{49.7} & \textbf{36.8} & 59.7 & \textbf{51.0} \\
    \bottomrule
    \end{tabular}
    }
    \caption{Answer accuracy averaged over all 4 test sets. {\em None} is the non-fine-tuned baseline, {\em Retrieve}, {\em Rerank} and {\em Hard} are fine-tuning strategies. Metrics: (1) \accu, accuracy on ungrounded instances, (2) \accg, accuracy on grounded instances, and (3) \acct, overall accuracy. Bold values indicate the highest per model and dataset. The LLMs fine-tuned on the \textit{Hard} dataset achieve statistically significant superior $acc$ in all test sets besides Llama-3.1-8B on PopQA where results are slightly lower than training on \textit{Rerank}, but in a non-statistically significant manner (Wilcoxon test with p-value < 0.01).}\label{tab:ft results}
\end{table}

\section{Conclusions} \label{sec:conclusions}
%

In this paper, we explored the topic of distracting passages in the context of RAG. We provided an algorithm to measure the distracting effect of a passage w.r.t. a query and LLM and demonstrated its robustness across LLM types and alternative implementations. We explored different ways to obtain distracting passages, going beyond the common approach of using standard retrieval. We showed that the combination of these methods produces more distracting passages; this allows the creation of more challenging and diverse datasets for RAG, and we demonstrated how they can be used to fine-tune LLMs to be more robust to distracting passages.
We note that this application represents one of the potentially many use cases, and we believe that the insights gained from our study of the distracting effect of passages will prove valuable for additional applications.

\section*{Limitations}



The generation categories discussed in Section \ref{sec:generation} do not necessarily capture the full range of distracting passage types. Expanding this taxonomy to account for additional rhetorical strategies or domain-specific use cases remains an open research question.

Moreover, our research primarily investigated the question-answering task, though the concept of distracting passages extends to various RAG use cases. Indeed, extending the study to additional tasks will provide a more complete picture, which we leave to future work. 

Finally, while we conducted our experiments on English-language benchmarks, the language-agnostic nature of our methodology suggests that the findings would likely generalize to other languages, though formal verification of this hypothesis remains to be carried out.

\section*{Acknowledgments}
This work was carried out while Florin Cuconasu was enrolled in the Italian National Doctorate on Artificial Intelligence run by the Sapienza University of Rome. This project has also been supported by PNRR MUR project PE0000013-FAIR.


\begin{thebibliography}{32}
\providecommand{\natexlab}[1]{#1}

\bibitem[{Abdumalikov et~al.(2024)Abdumalikov, Minervini, and Kementchedjhieva}]{abdumalikov2024answerability}
Rustam Abdumalikov, Pasquale Minervini, and Yova Kementchedjhieva. 2024.
\newblock Answerability in retrieval-augmented open-domain question answering.
\newblock \emph{arXiv preprint arXiv:2403.01461}.

\bibitem[{Asai et~al.(2024)Asai, Wu, Wang, Sil, and Hajishirzi}]{asaiself}
Akari Asai, Zeqiu Wu, Yizhong Wang, Avirup Sil, and Hannaneh Hajishirzi. 2024.
\newblock Self-{RAG}: Learning to retrieve, generate, and critique through self-reflection.
\newblock In \emph{The Twelfth International Conference on Learning Representations}.

\bibitem[{Basmov et~al.(2024)Basmov, Goldberg, and Tsarfaty}]{basmov2024llms}
Victoria Basmov, Yoav Goldberg, and Reut Tsarfaty. 2024.
\newblock {LLM}s' reading comprehension is affected by parametric knowledge and struggles with hypothetical statements.
\newblock \emph{arXiv preprint arXiv:2404.06283}.

\bibitem[{Berant et~al.(2013)Berant, Chou, Frostig, and Liang}]{berant-etal-2013-semantic}
Jonathan Berant, Andrew Chou, Roy Frostig, and Percy Liang. 2013.
\newblock \href {https://www.aclweb.org/anthology/D13-1160} {Semantic parsing on {F}reebase from question-answer pairs}.
\newblock In \emph{Proceedings of the 2013 Conference on Empirical Methods in Natural Language Processing}, pages 1533--1544, Seattle, Washington, USA. Association for Computational Linguistics.

\bibitem[{Chen et~al.(2017)Chen, Fisch, Weston, and Bordes}]{chen2017reading}
Danqi Chen, Adam Fisch, Jason Weston, and Antoine Bordes. 2017.
\newblock Reading wikipedia to answer open-domain questions.
\newblock In \emph{Proceedings of the 55th Annual Meeting of the Association for Computational Linguistics (Volume 1: Long Papers)}, pages 1870--1879.

\bibitem[{Chen et~al.(2024)Chen, Xiao, Zhang, Luo, Lian, and Liu}]{chen2024bge}
Jianlv Chen, Shitao Xiao, Peitian Zhang, Kun Luo, Defu Lian, and Zheng Liu. 2024.
\newblock \href {https://arxiv.org/abs/2402.03216} {{BGE} {M3}-embedding: Multi-lingual, multi-functionality, multi-granularity text embeddings through self-knowledge distillation}.
\newblock \emph{Preprint}, arXiv:2402.03216.

\bibitem[{Cuconasu et~al.(2024)Cuconasu, Trappolini, Siciliano, Filice, Campagnano, Maarek, Tonellotto, and Silvestri}]{cuconasu2024power}
Florin Cuconasu, Giovanni Trappolini, Federico Siciliano, Simone Filice, Cesare Campagnano, Yoelle Maarek, Nicola Tonellotto, and Fabrizio Silvestri. 2024.
\newblock The power of noise: Redefining retrieval for {RAG} systems.
\newblock In \emph{Proceedings of the 47th International ACM SIGIR Conference on Research and Development in Information Retrieval}, pages 719--729.

\bibitem[{Fan et~al.(2024)Fan, Ding, Ning, Wang, Li, Yin, Chua, and Li}]{fan2024survey}
Wenqi Fan, Yujuan Ding, Liangbo Ning, Shijie Wang, Hengyun Li, Dawei Yin, Tat-Seng Chua, and Qing Li. 2024.
\newblock A survey on {RAG} meeting {LLM}s: Towards retrieval-augmented large language models.
\newblock In \emph{Proceedings of the 30th ACM SIGKDD Conference on Knowledge Discovery and Data Mining}, pages 6491--6501.

\bibitem[{Gao et~al.(2023)Gao, Xiong, Gao, Jia, Pan, Bi, Dai, Sun, and Wang}]{gao2023retrieval}
Yunfan Gao, Yun Xiong, Xinyu Gao, Kangxiang Jia, Jinliu Pan, Yuxi Bi, Yi~Dai, Jiawei Sun, and Haofen Wang. 2023.
\newblock Retrieval-augmented generation for large language models: A survey.
\newblock \emph{arXiv preprint arXiv:2312.10997}.

\bibitem[{Grattafiori et~al.(2024)Grattafiori, Dubey, Jauhri, Pandey, Kadian, Al-Dahle, Letman, Mathur, Schelten, Vaughan, Yang, Fan, Goyal, Hartshorn, Yang, Mitra, Sravankumar, Korenev, Hinsvark, Rao, Zhang, Rodriguez, Gregerson, Spataru, Roziere, Biron, Tang, Chern, Caucheteux, Nayak, Bi, Marra, McConnell, Keller, Touret, Wu, Wong, Ferrer, Nikolaidis, Allonsius, Song, Pintz, Livshits, Wyatt, Esiobu, Choudhary, Mahajan, Garcia-Olano, Perino, Hupkes, Lakomkin, AlBadawy, Lobanova, Dinan, Smith, Radenovic, Guzmán, Zhang, Synnaeve, Lee, Anderson, Thattai, Nail, Mialon, Pang, Cucurell, Nguyen, Korevaar, Xu, Touvron, Zarov, Ibarra, Kloumann, Misra, Evtimov, Zhang, Copet, Lee, Geffert, Vranes, Park, Mahadeokar, Shah, van~der Linde, Billock, Hong, Lee, Fu, Chi, Huang, Liu, Wang, Yu, Bitton, Spisak, Park, Rocca, Johnstun, Saxe, Jia, Alwala, Prasad, Upasani, Plawiak, Li, Heafield, Stone, El-Arini, Iyer, Malik, Chiu, Bhalla, Lakhotia, Rantala-Yeary, van~der Maaten, Chen, Tan, Jenkins, Martin, Madaan, Malo, Blecher,
  Landzaat, de~Oliveira, Muzzi, Pasupuleti, Singh, Paluri, Kardas, Tsimpoukelli, Oldham, Rita, Pavlova, Kambadur, Lewis, Si, Singh, Hassan, Goyal, Torabi, Bashlykov, Bogoychev, Chatterji, Zhang, Duchenne, Çelebi, Alrassy, Zhang, Li, Vasic, Weng, Bhargava, Dubal, Krishnan, Koura, Xu, He, Dong, Srinivasan, Ganapathy, Calderer, Cabral, Stojnic, Raileanu, Maheswari, Girdhar, Patel, Sauvestre, Polidoro, Sumbaly, Taylor, Silva, Hou, Wang, Hosseini, Chennabasappa, Singh, Bell, Kim, Edunov, Nie, Narang, Raparthy, Shen, Wan, Bhosale, Zhang, Vandenhende, Batra, Whitman, Sootla, Collot, Gururangan, Borodinsky, Herman, Fowler, Sheasha, Georgiou, Scialom, Speckbacher, Mihaylov, Xiao, Karn, Goswami, Gupta, Ramanathan, Kerkez, Gonguet, Do, Vogeti, Albiero, Petrovic, Chu, Xiong, Fu, Meers, Martinet, Wang, Wang, Tan, Xia, Xie, Jia, Wang, Goldschlag, Gaur, Babaei, Wen, Song, Zhang, Li, Mao, Coudert, Yan, Chen, Papakipos, Singh, Srivastava, Jain, Kelsey, Shajnfeld, Gangidi, Victoria, Goldstand, Menon, Sharma, Boesenberg,
  Baevski, Feinstein, Kallet, Sangani, Teo, Yunus, Lupu, Alvarado, Caples, Gu, Ho, Poulton, Ryan, Ramchandani, Dong, Franco, Goyal, Saraf, Chowdhury, Gabriel, Bharambe, Eisenman, Yazdan, James, Maurer, Leonhardi, Huang, Loyd, Paola, Paranjape, Liu, Wu, Ni, Hancock, Wasti, Spence, Stojkovic, Gamido, Montalvo, Parker, Burton, Mejia, Liu, Wang, Kim, Zhou, Hu, Chu, Cai, Tindal, Feichtenhofer, Gao, Civin, Beaty, Kreymer, Li, Adkins, Xu, Testuggine, David, Parikh, Liskovich, Foss, Wang, Le, Holland, Dowling, Jamil, Montgomery, Presani, Hahn, Wood, Le, Brinkman, Arcaute, Dunbar, Smothers, Sun, Kreuk, Tian, Kokkinos, Ozgenel, Caggioni, Kanayet, Seide, Florez, Schwarz, Badeer, Swee, Halpern, Herman, Sizov, Guangyi, Zhang, Lakshminarayanan, Inan, Shojanazeri, Zou, Wang, Zha, Habeeb, Rudolph, Suk, Aspegren, Goldman, Zhan, Damlaj, Molybog, Tufanov, Leontiadis, Veliche, Gat, Weissman, Geboski, Kohli, Lam, Asher, Gaya, Marcus, Tang, Chan, Zhen, Reizenstein, Teboul, Zhong, Jin, Yang, Cummings, Carvill, Shepard, McPhie,
  Torres, Ginsburg, Wang, Wu, U, Saxena, Khandelwal, Zand, Matosich, Veeraraghavan, Michelena, Li, Jagadeesh, Huang, Chawla, Huang, Chen, Garg, A, Silva, Bell, Zhang, Guo, Yu, Moshkovich, Wehrstedt, Khabsa, Avalani, Bhatt, Mankus, Hasson, Lennie, Reso, Groshev, Naumov, Lathi, Keneally, Liu, Seltzer, Valko, Restrepo, Patel, Vyatskov, Samvelyan, Clark, Macey, Wang, Hermoso, Metanat, Rastegari, Bansal, Santhanam, Parks, White, Bawa, Singhal, Egebo, Usunier, Mehta, Laptev, Dong, Cheng, Chernoguz, Hart, Salpekar, Kalinli, Kent, Parekh, Saab, Balaji, Rittner, Bontrager, Roux, Dollar, Zvyagina, Ratanchandani, Yuvraj, Liang, Alao, Rodriguez, Ayub, Murthy, Nayani, Mitra, Parthasarathy, Li, Hogan, Battey, Wang, Howes, Rinott, Mehta, Siby, Bondu, Datta, Chugh, Hunt, Dhillon, Sidorov, Pan, Mahajan, Verma, Yamamoto, Ramaswamy, Lindsay, Lindsay, Feng, Lin, Zha, Patil, Shankar, Zhang, Zhang, Wang, Agarwal, Sajuyigbe, Chintala, Max, Chen, Kehoe, Satterfield, Govindaprasad, Gupta, Deng, Cho, Virk, Subramanian, Choudhury,
  Goldman, Remez, Glaser, Best, Koehler, Robinson, Li, Zhang, Matthews, Chou, Shaked, Vontimitta, Ajayi, Montanez, Mohan, Kumar, Mangla, Ionescu, Poenaru, Mihailescu, Ivanov, Li, Wang, Jiang, Bouaziz, Constable, Tang, Wu, Wang, Wu, Gao, Kleinman, Chen, Hu, Jia, Qi, Li, Zhang, Zhang, Adi, Nam, Yu, Wang, Zhao, Hao, Qian, Li, He, Rait, DeVito, Rosnbrick, Wen, Yang, Zhao, and Ma}]{grattafiori2024llama3herdmodels}
Aaron Grattafiori, Abhimanyu Dubey, Abhinav Jauhri, Abhinav Pandey, Abhishek Kadian, Ahmad Al-Dahle, Aiesha Letman, Akhil Mathur, Alan Schelten, Alex Vaughan, Amy Yang, Angela Fan, Anirudh Goyal, Anthony Hartshorn, Aobo Yang, Archi Mitra, Archie Sravankumar, Artem Korenev, Arthur Hinsvark, and 542 others. 2024.
\newblock \href {https://arxiv.org/abs/2407.21783} {The llama 3 herd of models}.
\newblock \emph{Preprint}, arXiv:2407.21783.

\bibitem[{Honovich et~al.(2022)Honovich, Aharoni, Herzig, Taitelbaum, Kukliansy, Cohen, Scialom, Szpektor, Hassidim, and Matias}]{honovich2022true}
Or~Honovich, Roee Aharoni, Jonathan Herzig, Hagai Taitelbaum, Doron Kukliansy, Vered Cohen, Thomas Scialom, Idan Szpektor, Avinatan Hassidim, and Yossi Matias. 2022.
\newblock True: Re-evaluating factual consistency evaluation.
\newblock In \emph{Proceedings of the 2022 Conference of the North American Chapter of the Association for Computational Linguistics: Human Language Technologies}, pages 3905--3920.

\bibitem[{Hu et~al.(2021)Hu, Shen, Wallis, Allen-Zhu, Li, Wang, Wang, and Chen}]{hu2021lora}
Edward~J Hu, Yelong Shen, Phillip Wallis, Zeyuan Allen-Zhu, Yuanzhi Li, Shean Wang, Lu~Wang, and Weizhu Chen. 2021.
\newblock Lora: Low-rank adaptation of large language models.
\newblock \emph{arXiv preprint arXiv:2106.09685}.

\bibitem[{Jin et~al.(2024)Jin, Yoon, Han, and Arik}]{jin2024long}
Bowen Jin, Jinsung Yoon, Jiawei Han, and Sercan~O Arik. 2024.
\newblock Long-context {LLM}s meet {RAG}: Overcoming challenges for long inputs in {RAG}.
\newblock \emph{arXiv preprint arXiv:2410.05983}.

\bibitem[{Joshi et~al.(2017)Joshi, Choi, Weld, and Zettlemoyer}]{joshi2017triviaqa}
Mandar Joshi, Eunsol Choi, Daniel~S Weld, and Luke Zettlemoyer. 2017.
\newblock {TriviaQA}: A large scale distantly supervised challenge dataset for reading comprehension.
\newblock In \emph{Proceedings of the 55th Annual Meeting of the Association for Computational Linguistics (Volume 1: Long Papers)}, pages 1601--1611.

\bibitem[{Kwiatkowski et~al.(2019)Kwiatkowski, Palomaki, Redfield, Collins, Parikh, Alberti, Epstein, Polosukhin, Devlin, Lee et~al.}]{kwiatkowski2019natural}
Tom Kwiatkowski, Jennimaria Palomaki, Olivia Redfield, Michael Collins, Ankur Parikh, Chris Alberti, Danielle Epstein, Illia Polosukhin, Jacob Devlin, Kenton Lee, and 1 others. 2019.
\newblock Natural questions: a benchmark for question answering research.
\newblock \emph{Transactions of the Association for Computational Linguistics}, 7:453--466.

\bibitem[{Li et~al.(2023)Li, Rawat, Zaheer, Wang, Lukasik, Veit, Yu, and Kumar}]{li2023large}
Daliang Li, Ankit~Singh Rawat, Manzil Zaheer, Xin Wang, Michal Lukasik, Andreas Veit, Felix Yu, and Sanjiv Kumar. 2023.
\newblock Large language models with controllable working memory.
\newblock In \emph{Findings of the Association for Computational Linguistics: ACL 2023}, pages 1774--1793.

\bibitem[{Lin et~al.(2024)Lin, Chen, Chen, Shi, Lomeli, James, Rodriguez, Kahn, Szilvasy, Lewis, Zettlemoyer, and Yih}]{linra2024}
Xi~Victoria Lin, Xilun Chen, Mingda Chen, Weijia Shi, Maria Lomeli, Richard James, Pedro Rodriguez, Jacob Kahn, Gergely Szilvasy, Mike Lewis, Luke Zettlemoyer, and Scott Yih. 2024.
\newblock {RA-DIT}: Retrieval-augmented dual instruction tuning.
\newblock In \emph{The Twelfth International Conference on Learning Representations}.

\bibitem[{Liu et~al.(2023)Liu, Lin, Hewitt, Paranjape, Bevilacqua, Petroni, and Liang}]{liu2023lostmiddlelanguagemodels}
Nelson~F. Liu, Kevin Lin, John Hewitt, Ashwin Paranjape, Michele Bevilacqua, Fabio Petroni, and Percy Liang. 2023.
\newblock \href {https://arxiv.org/abs/2307.03172} {Lost in the middle: How language models use long contexts}.
\newblock \emph{Preprint}, arXiv:2307.03172.

\bibitem[{Luo et~al.(2023)Luo, Chuang, Gong, Zhang, Kim, Wu, Fox, Meng, and Glass}]{luo2023sail}
Hongyin Luo, Yung-Sung Chuang, Yuan Gong, Tianhua Zhang, Yoon Kim, Xixin Wu, Danny Fox, Helen Meng, and James Glass. 2023.
\newblock Sail: Search-augmented instruction learning.
\newblock \emph{arXiv preprint arXiv:2305.15225}.

\bibitem[{Mallen et~al.(2023)Mallen, Asai, Zhong, Das, Khashabi, and Hajishirzi}]{mallen2023not}
Alex~Troy Mallen, Akari Asai, Victor Zhong, Rajarshi Das, Daniel Khashabi, and Hannaneh Hajishirzi. 2023.
\newblock When not to trust language models: Investigating effectiveness of parametric and non-parametric memories.
\newblock In \emph{The 61st Annual Meeting Of The Association For Computational Linguistics}.

\bibitem[{Moreira et~al.(2024)Moreira, Osmulski, Xu, Ak, Schifferer, and Oldridge}]{moreira2024nv}
Gabriel de Souza~P Moreira, Radek Osmulski, Mengyao Xu, Ronay Ak, Benedikt Schifferer, and Even Oldridge. 2024.
\newblock {NV-Retriever}: Improving text embedding models with effective hard-negative mining.
\newblock \emph{arXiv preprint arXiv:2407.15831}.

\bibitem[{Petroni et~al.(2021)Petroni, Piktus, Fan, Lewis, Yazdani, De~Cao, Thorne, Jernite, Karpukhin, Maillard, Plachouras, Rocktäschel, and Riedel}]{petroni2021kilt}
Fabio Petroni, Aleksandra Piktus, Angela Fan, Patrick Lewis, Majid Yazdani, Nicola De~Cao, James Thorne, Yacine Jernite, Vladimir Karpukhin, Jean Maillard, Vassilis Plachouras, Tim Rocktäschel, and Sebastian Riedel. 2021.
\newblock {KILT}: a benchmark for knowledge intensive language tasks.
\newblock In \emph{Proceedings of the 2021 Conference of the North American Chapter of the Association for Computational Linguistics: Human Language Technologies}, pages 2523--2544.

\bibitem[{Team(2024)}]{Falcon3}
Falcon-LLM Team. 2024.
\newblock \href {https://huggingface.co/blog/falcon3} {The falcon 3 family of open models}.

\bibitem[{Wang et~al.(2022)Wang, Yang, Huang, Jiao, Yang, Jiang, Majumder, and Wei}]{wang2022text}
Liang Wang, Nan Yang, Xiaolong Huang, Binxing Jiao, Linjun Yang, Daxin Jiang, Rangan Majumder, and Furu Wei. 2022.
\newblock Text embeddings by weakly-supervised contrastive pre-training.
\newblock \emph{arXiv preprint arXiv:2212.03533}.

\bibitem[{Wei et~al.(2024)Wei, Chen, and Meng}]{weiinstructrag2024}
Zhepei Wei, Wei-Lin Chen, and Yu~Meng. 2024.
\newblock Instructrag: Instructing retrieval augmented generation via self-synthesized rationales.
\newblock In \emph{Adaptive Foundation Models: Evolving AI for Personalized and Efficient Learning}.

\bibitem[{Weller et~al.(2024)Weller, Van~Durme, Lawrie, Paranjape, Zhang, and Hessel}]{weller2024promptriever}
Orion Weller, Benjamin Van~Durme, Dawn Lawrie, Ashwin Paranjape, Yuhao Zhang, and Jack Hessel. 2024.
\newblock Promptriever: Instruction-trained retrievers can be prompted like language models.
\newblock \emph{arXiv preprint arXiv:2409.11136}.

\bibitem[{Wolf et~al.(2020)Wolf, Debut, Sanh, Chaumond, Delangue, Moi, Cistac, Rault, Louf, Funtowicz, Davison, Shleifer, von Platen, Ma, Jernite, Plu, Xu, Scao, Gugger, Drame, Lhoest, and Rush}]{wolf2020transformers}
Thomas Wolf, Lysandre Debut, Victor Sanh, Julien Chaumond, Clement Delangue, Anthony Moi, Pierric Cistac, Tim Rault, R{\'{e}}mi Louf, Morgan Funtowicz, Joe Davison, Sam Shleifer, Patrick von Platen, Clara Ma, Yacine Jernite, Julien Plu, Canwen Xu, Teven~Le Scao, Sylvain Gugger, and 3 others. 2020.
\newblock \href {https://doi.org/10.18653/V1/2020.EMNLP-DEMOS.6} {Transformers: State-of-the-art natural language processing}.
\newblock In \emph{Proceedings of the 2020 Conference on Empirical Methods in Natural Language Processing: System Demonstrations, {EMNLP} 2020 - Demos, Online, November 16-20, 2020}, pages 38--45. Association for Computational Linguistics.

\bibitem[{Xiong et~al.(2021)Xiong, Xiong, Li, Tang, Liu, Bennett, Ahmed, and Overwijk}]{xiongapproximate2021}
Lee Xiong, Chenyan Xiong, Ye~Li, Kwok-Fung Tang, Jialin Liu, Paul~N Bennett, Junaid Ahmed, and Arnold Overwijk. 2021.
\newblock Approximate nearest neighbor negative contrastive learning for dense text retrieval.
\newblock In \emph{International Conference on Learning Representations}.

\bibitem[{Yan et~al.(2024)Yan, Gu, Zhu, and Ling}]{yan2024corrective}
Shi-Qi Yan, Jia-Chen Gu, Yun Zhu, and Zhen-Hua Ling. 2024.
\newblock Corrective retrieval augmented generation.
\newblock \emph{arXiv preprint arXiv:2401.15884}.

\bibitem[{Yang et~al.(2025)Yang, Yang, Zhang, Hui, Zheng, Yu, Li, Liu, Huang, Wei, Lin, Yang, Tu, Zhang, Yang, Yang, Zhou, Lin, Dang, Lu, Bao, Yang, Yu, Li, Xue, Zhang, Zhu, Men, Lin, Li, Tang, Xia, Ren, Ren, Fan, Su, Zhang, Wan, Liu, Cui, Zhang, and Qiu}]{qwen2025qwen25technicalreport}
An~Yang, Baosong Yang, Beichen Zhang, Binyuan Hui, Bo~Zheng, Bowen Yu, Chengyuan Li, Dayiheng Liu, Fei Huang, Haoran Wei, Huan Lin, Jian Yang, Jianhong Tu, Jianwei Zhang, Jianxin Yang, Jiaxi Yang, Jingren Zhou, Junyang Lin, Kai Dang, and 23 others. 2025.
\newblock \href {https://arxiv.org/abs/2412.15115} {Qwen2.5 technical report}.
\newblock \emph{Preprint}, arXiv:2412.15115.

\bibitem[{Yoran et~al.(2024)Yoran, Wolfson, Ram, and Berant}]{yoran2024making}
Ori Yoran, Tomer Wolfson, Ori Ram, and Jonathan Berant. 2024.
\newblock Making retrieval-augmented language models robust to irrelevant context.
\newblock In \emph{The Twelfth International Conference on Learning Representations}.

\bibitem[{Yu et~al.(2023)Yu, Zhang, Pan, Ma, Wang, and Yu}]{yu2023chain}
Wenhao Yu, Hongming Zhang, Xiaoman Pan, Kaixin Ma, Hongwei Wang, and Dong Yu. 2023.
\newblock Chain-of-note: Enhancing robustness in retrieval-augmented language models.
\newblock \emph{arXiv preprint arXiv:2311.09210}.

\end{thebibliography}

\appendix

\section{Additional Details on Distracting Effect}
\label{sec:appendix_details_distracting}
In Section \ref{sec:score_candidates}, we introduce a method to quantify how distracting irrelevant passages are for LLMs using Equation \ref{eq:distracting_effect}. To compute this distracting effect for a given query $q$ and passage $p$, we follow Algorithm \ref{algo} which calculates $DE_q(p)$. The process begins by constructing a prompt using the template shown in Figure \ref{fig:prompt dist test}, where we explicitly include ``NO-RESPONSE'' as an answer (in the algorithm we refer to it as \texttt{target}). This prompt instructs the LLM to respond with ``NO-RESPONSE'' when the context contains no relevant answer.
We then measure the LLM's likelihood of generating ``NO-RESPONSE'' by examining the probability of its first token (that is $p^{\text{LLM}}(\text{NO-RESPONSE}|q,p)$ in Equation \ref{eq:distracting_effect}). This probability serves as a confidence measure: when it is high (making $DE_q(p)$ close to 0), the LLM is likely to abstain from answering. Conversely, when the probability is low (making $DE_q(p)$ close to 1), the LLM is more inclined to generate an answer based on the passage, indicating that the passage is distracting the LLM.

\begin{figure*}
\begin{mdframed}[font=\footnotesize]
\begin{Verbatim}[breaklines=true, breaksymbol=]
You are given a question and you must respond based on the provided documents. Respond directly without providing any premise or explanation. If none of the documents contain the answer, please respond with NO-RESPONSE. Do not try to respond based on your own knowledge.

Documents:
<document>

Question: 
<question>

Answer: NO-RESPONSE
\end{Verbatim}
\end{mdframed}
\vspace{-0.4cm}
\caption{Prompt for evaluating the distracting effect of a passage.} \label{fig:prompt dist test}
\end{figure*}

\begin{algorithm*}[t!]
\caption{Computing the Distracting Effect}
\label{algo}
\textbf{Input:} Query $q$, passage $p$, LLM $\mathcal{M}$, tokenizer $\tau$\\
\textbf{Output:} Distracting effect $DE_q(p)$
\begin{algorithmic}[1]
\State $\texttt{prompt} \gets \texttt{create\_prompt}(q, p)$ \Comment{Create the prompt using template in Figure \ref{fig:prompt dist test}}

\State $\texttt{tokens} \gets \tau(\texttt{prompt})$
\State $\texttt{target\_pos} \gets \texttt{get\_position}(\texttt{prompt}, \text{"NO-RESPONSE"})$ \Comment{Starting pos in the tokenized prompt}
\State $\texttt{target\_token} \gets \tau(\text{"NO-RESPONSE"})[0]$ \Comment{Get first target token}
\State $\texttt{logits} \gets \mathcal{M}(\texttt{tokens})$
\State $\texttt{probs} \gets \texttt{softmax}(\texttt{logits})$
\State $\texttt{prob} \gets \texttt{probs}[\texttt{target\_pos} - 1, \texttt{target\_token}]$ \Comment{Generation probability of the first target token}
\State \Return $1 - \texttt{prob}$
\end{algorithmic}
\end{algorithm*}

\subsection{Distracting Effect on Other LLMs and Datasets}
\label{sec:appendix_distracting_others}

\begin{figure*}[t!]
    \centering
    \includegraphics[width=\linewidth]{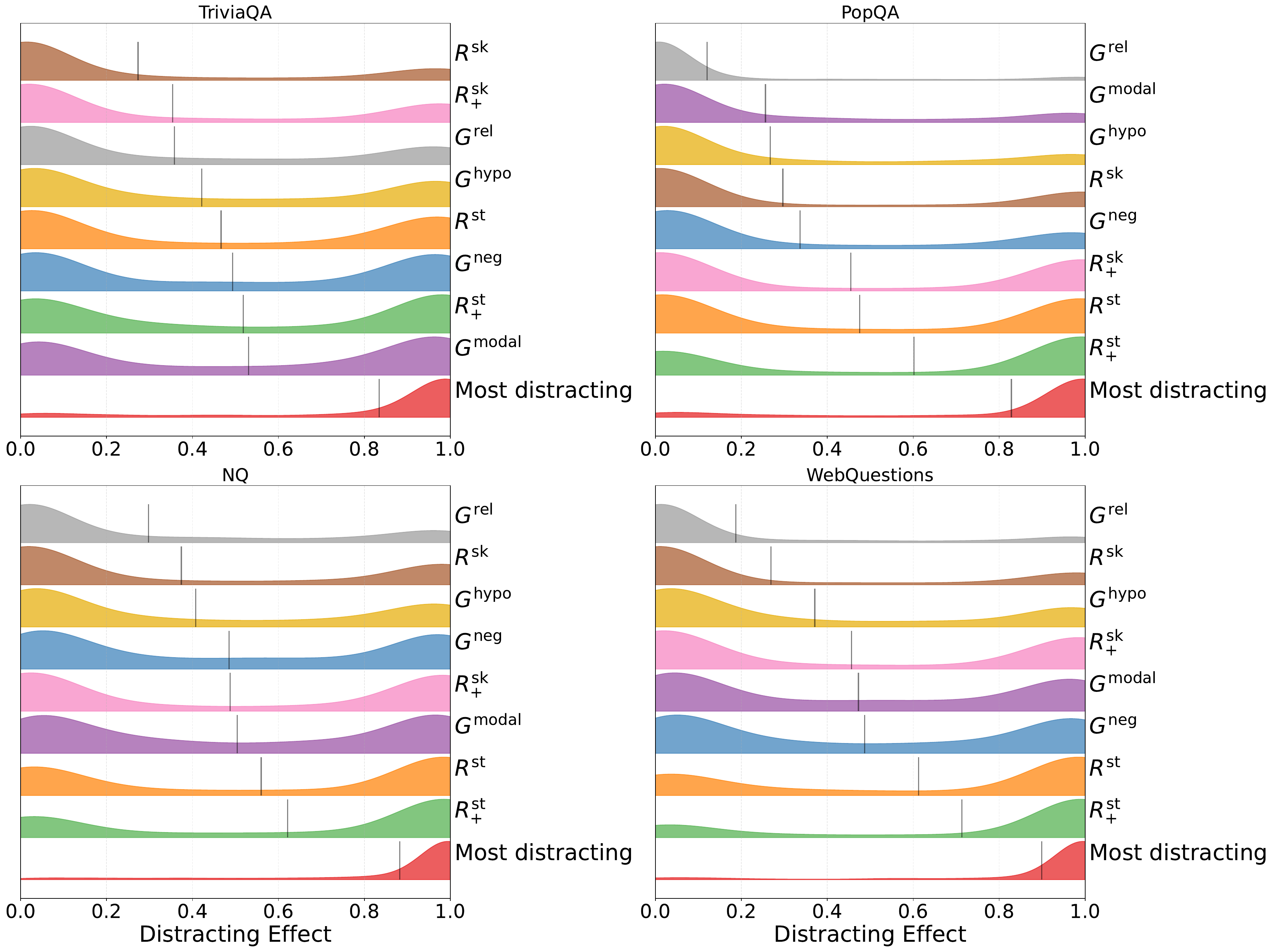}
    \caption{Distribution of distracting effect for passages obtained through different methods, as measured by Llama-3.1-8B on all datasets. Methods are ordered by their mean distracting effect (shown by vertical black lines), with higher means indicating a greater ability to distract the model.}
    \label{fig:llama3_8B_distracting_distribution}
\end{figure*}

\begin{figure*}[t!]
    \centering
    \includegraphics[width=\linewidth]{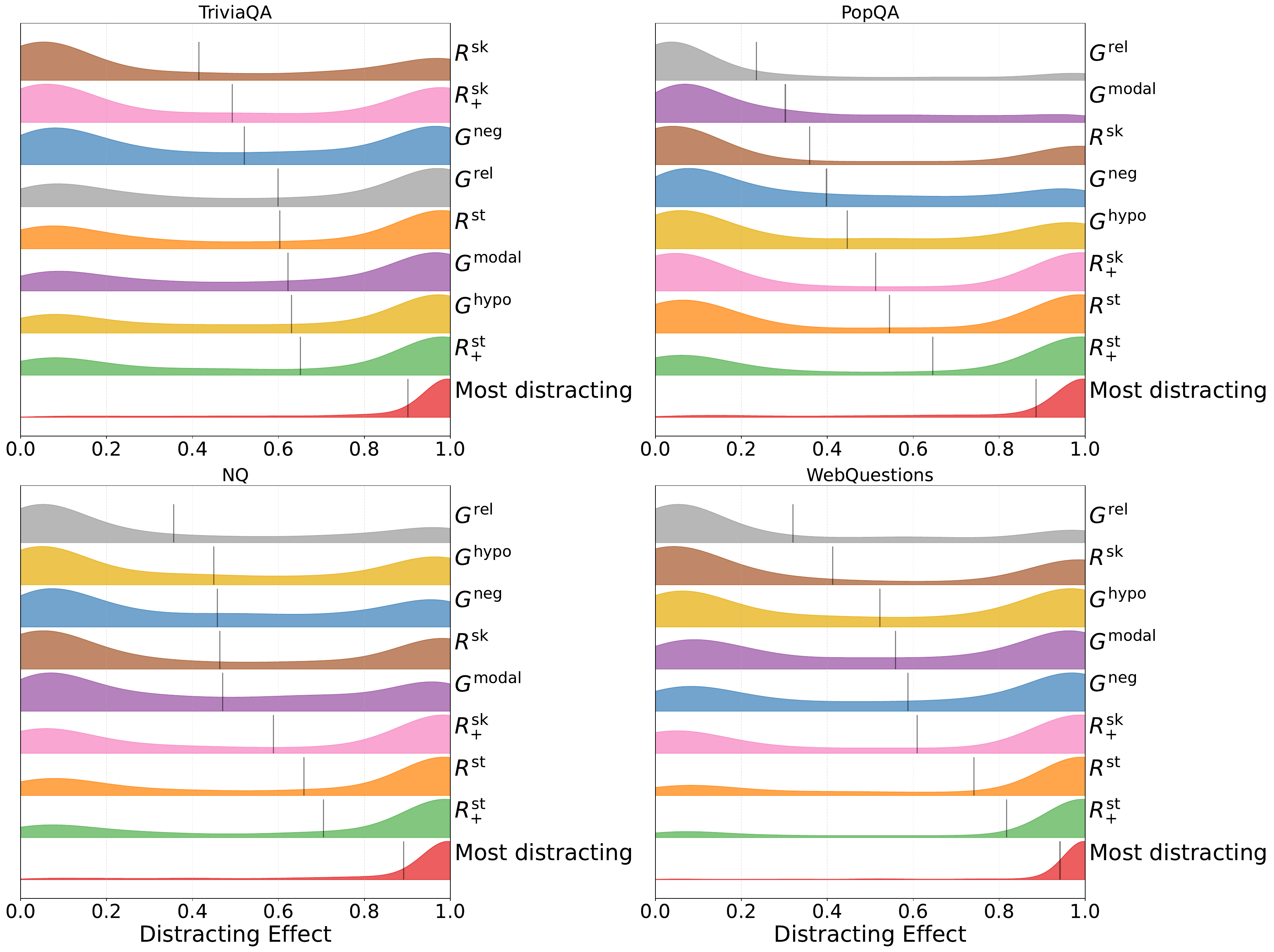}
    \caption{Distribution of distracting effect for passages obtained through different methods, as measured by Llama-3.2-3B on all datasets. Methods are ordered by their mean distracting effect (shown by vertical black lines), with higher means indicating a greater ability to distract the model.}
\label{fig:llama3_3B_distracting_distribution}
\end{figure*}

\begin{figure*}[t!]
    \centering
    \includegraphics[width=\linewidth]{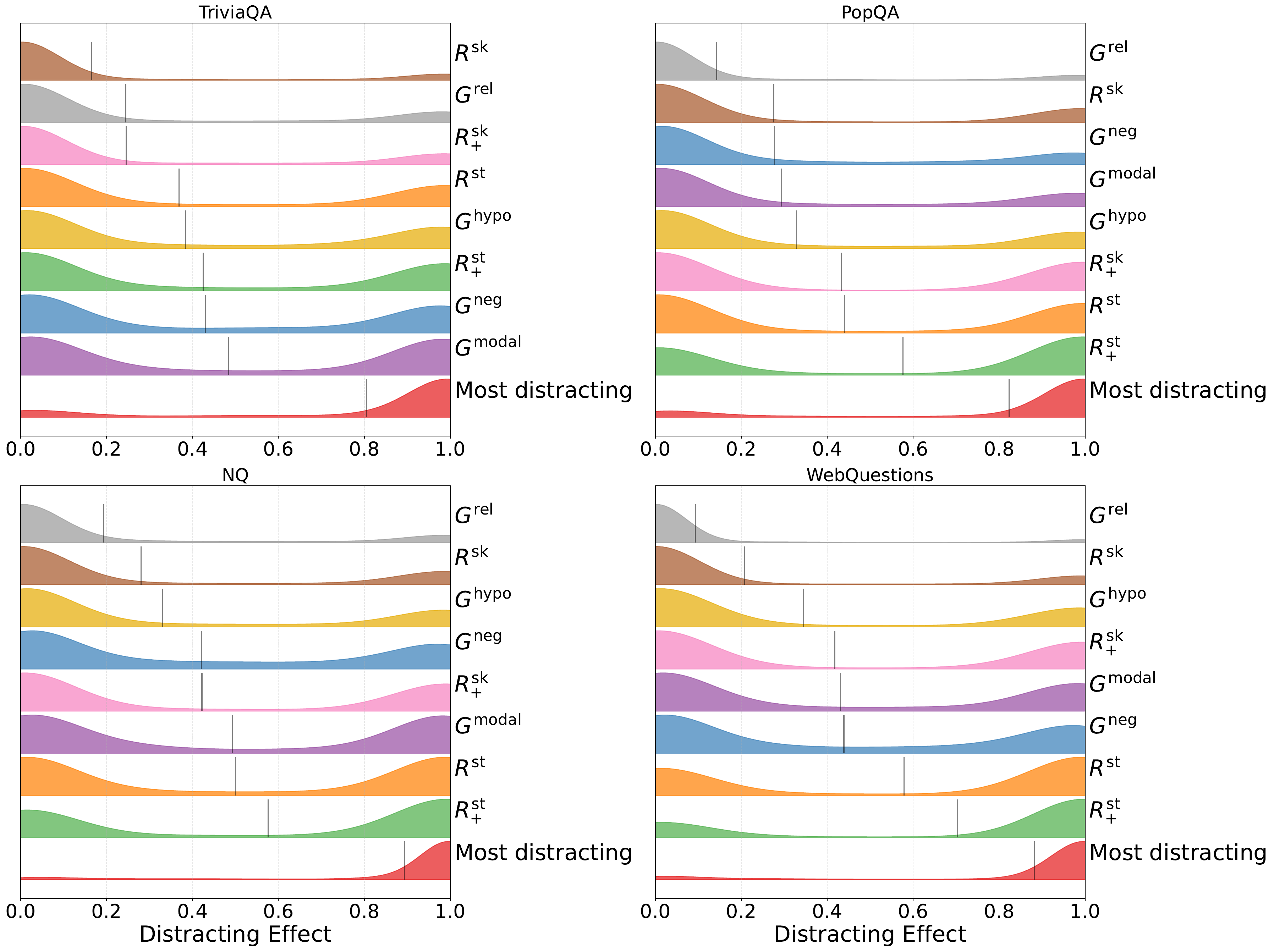}
    \caption{Distribution of distracting effect for passages obtained through different methods, as measured by Llama-3.3-70B on all datasets. Methods are ordered by their mean distracting effect (shown by vertical black lines), with higher means indicating a greater ability to distract the model.}
\label{fig:llama3_70B_distracting_distribution}
\end{figure*}

\begin{figure*}[t!]
    \centering
    \includegraphics[width=\linewidth]{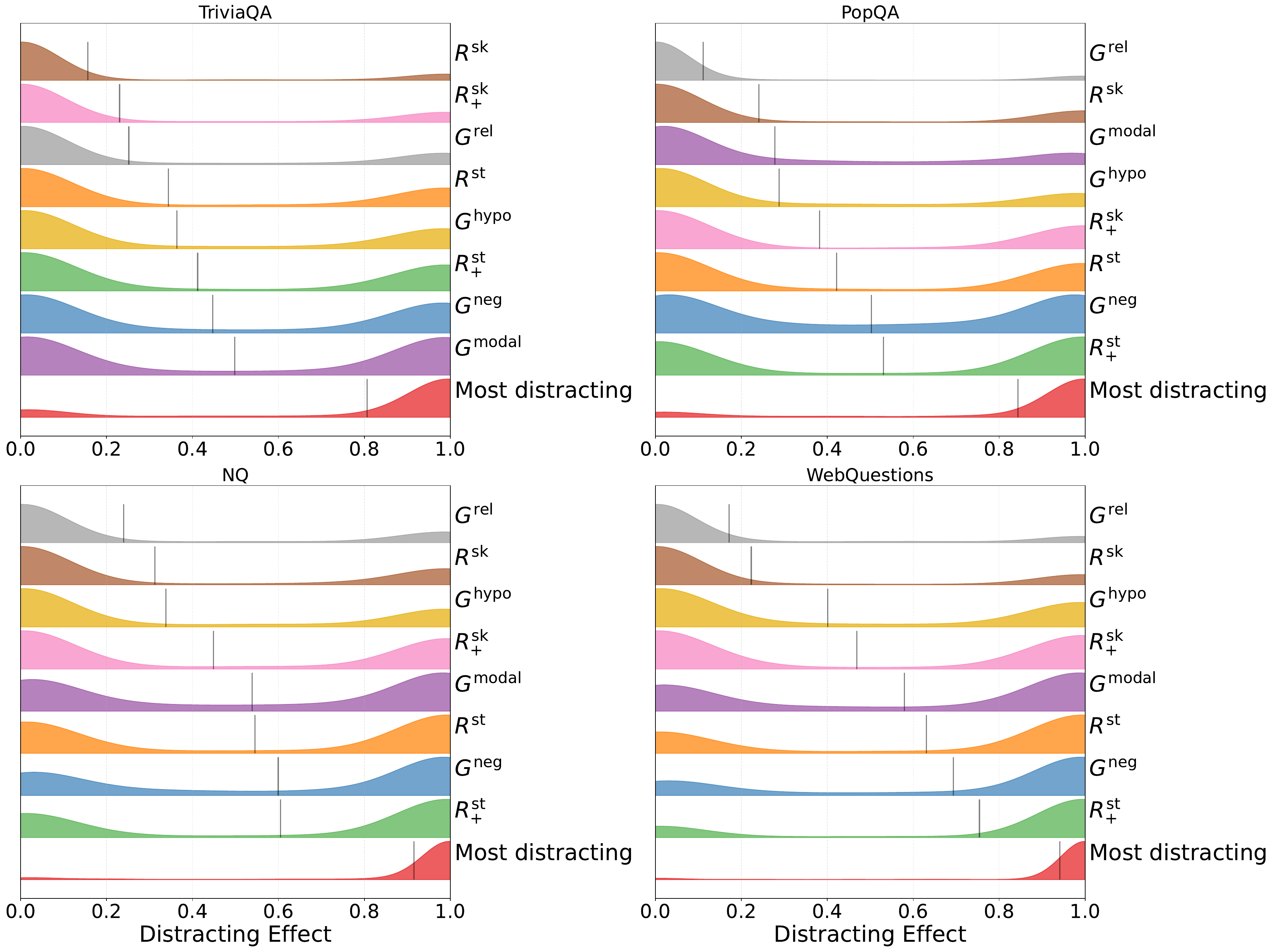}
    \caption{Distribution of distracting effect for passages obtained through different methods, as measured by Qwen-2.5-3B on all datasets. Methods are ordered by their mean distracting effect (shown by vertical black lines), with higher means indicating a greater ability to distract the model.}
\label{fig:qwen_3B_distracting_distribution}
\end{figure*}

\begin{figure*}[t]
    \centering
    \includegraphics[width=\linewidth]{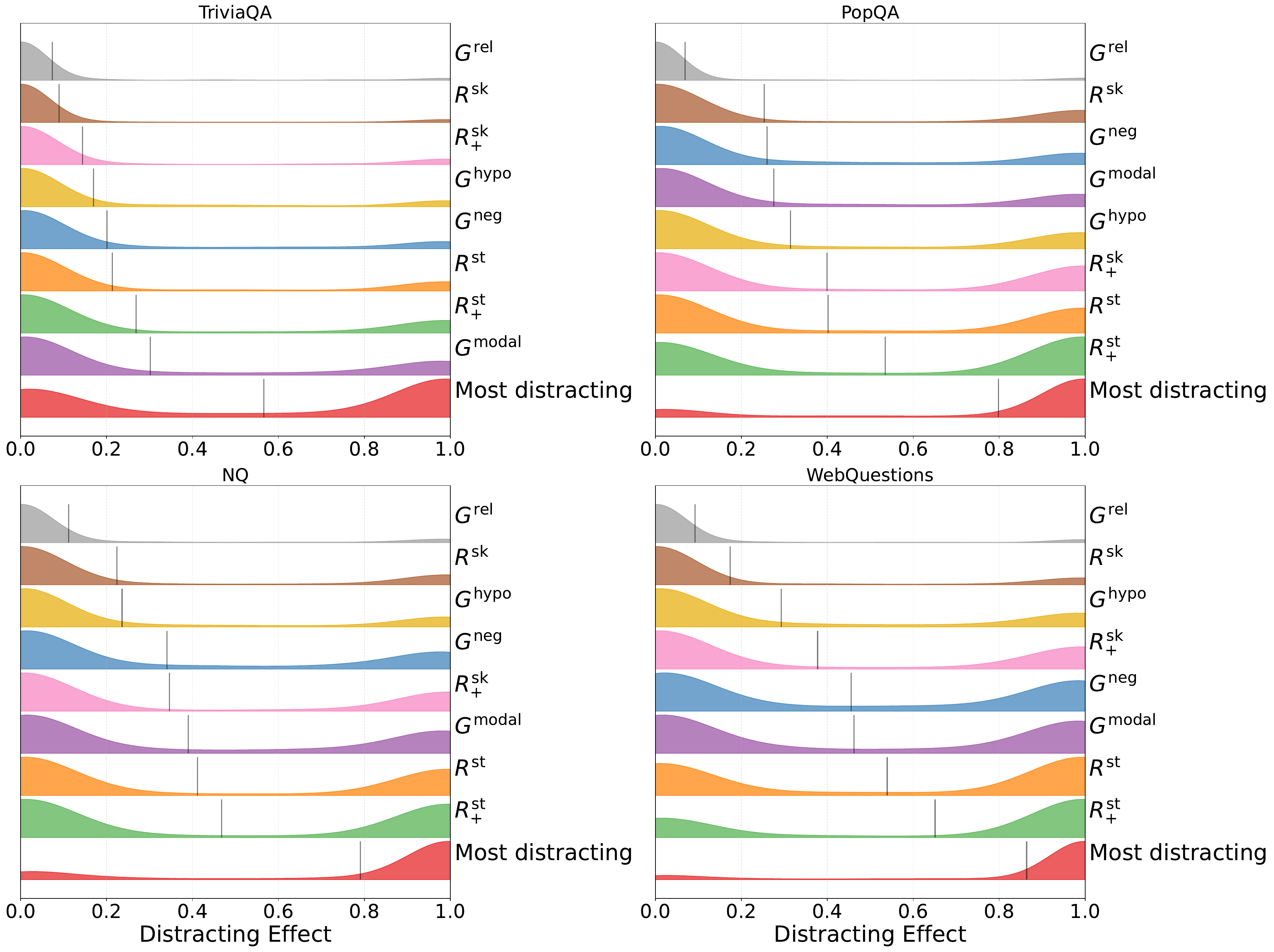}
    \caption{Distribution of distracting effect for passages obtained through different methods, as measured by Qwen-2.5-7B on all datasets. Methods are ordered by their mean distracting effect (shown by vertical black lines), with higher means indicating a greater ability to distract the model.}
\label{fig:qwen_7B_distracting_distribution}
\end{figure*}

\begin{figure*}[t]
    \centering
    \includegraphics[width=\linewidth]{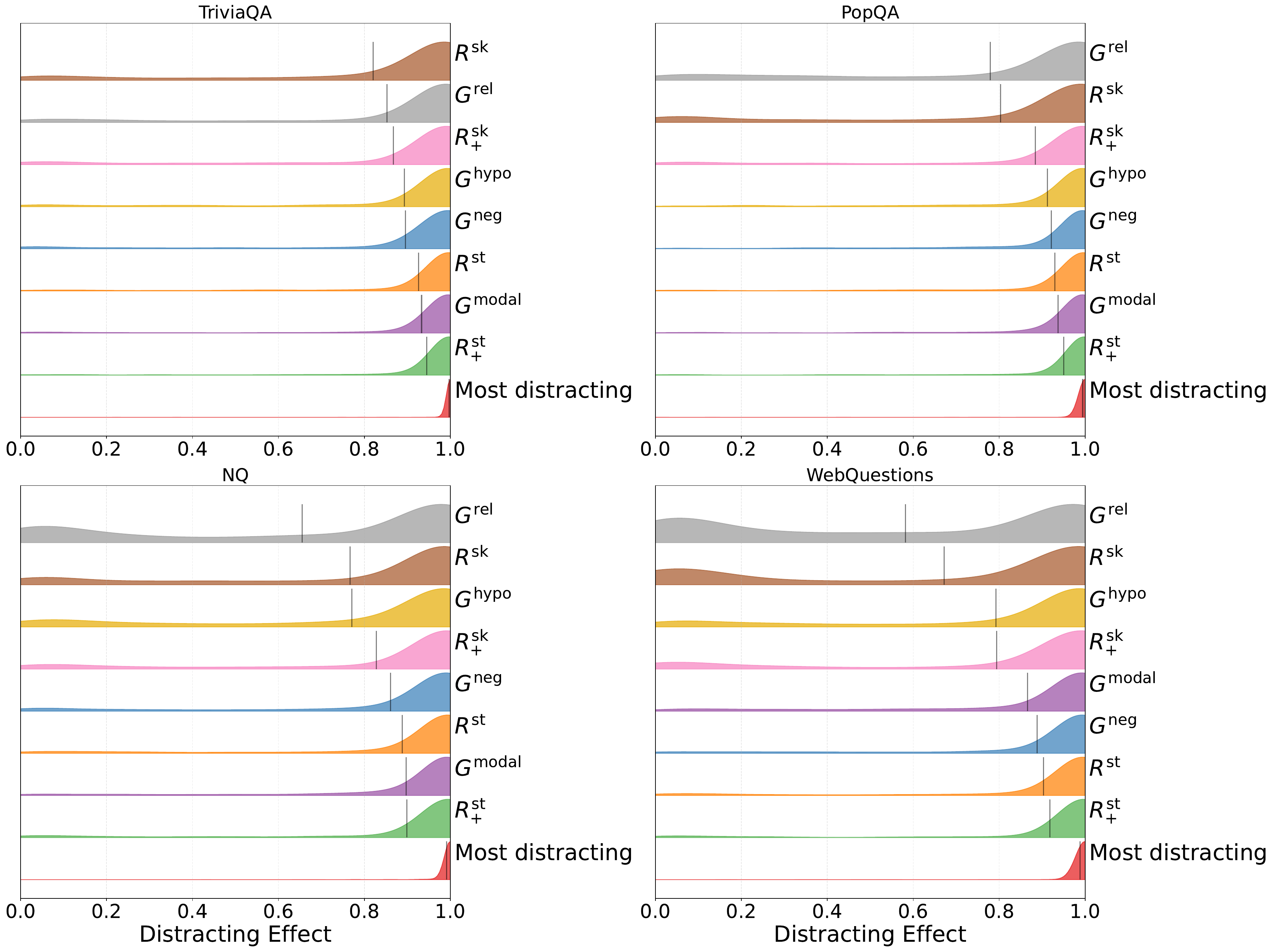}
    \caption{Distribution of distracting effect for passages obtained through different methods, as measured by Falcon-3-3B on all datasets. Methods are ordered by their mean distracting effect (shown by vertical black lines), with higher means indicating a greater ability to distract the model.}
\label{fig:falcon_3B_distracting_distribution}
\end{figure*}

\begin{figure*}[t]
    \centering
    \includegraphics[width=\linewidth]{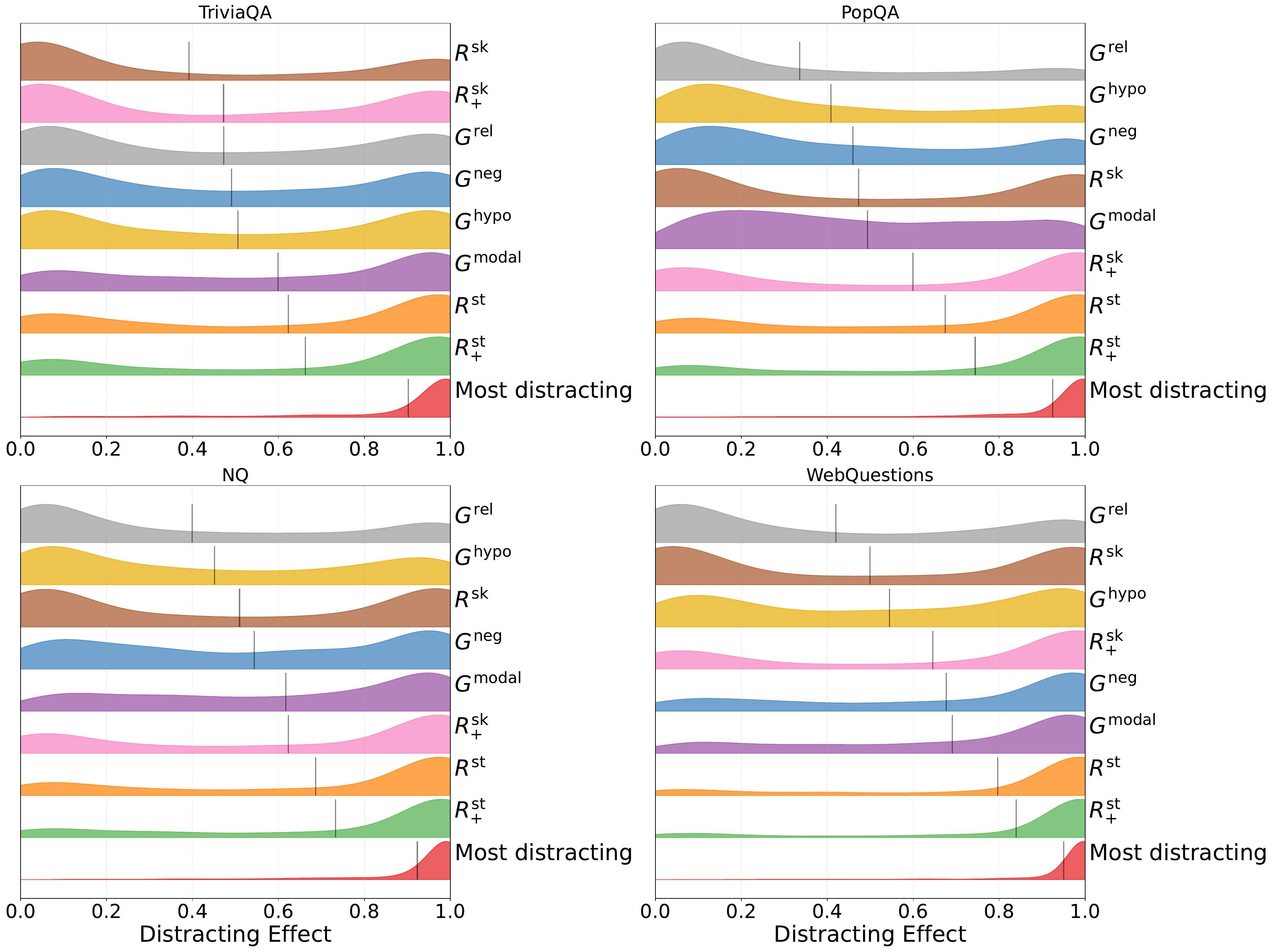}
    \caption{Distribution of distracting effect for passages obtained through different methods, as measured by Falcon-3-7B on all datasets. Methods are ordered by their mean distracting effect (shown by vertical black lines), with higher means indicating a greater ability to distract the model.}
\label{fig:falcon_7B_distracting_distribution}
\end{figure*}

In this section, we extend our analysis of distracting effects beyond Llama-3.1-8B on NQ and WebQuestions to other LLMs and datasets. As anticipated in Section~\ref{sec:compare_distracting_effects}, the probability distributions of distracting effects for TriviaQA and PopQA follow similar patterns to those observed for NQ and WebQuestions (Figure~\ref{fig:llama3_8B_distracting_distribution}). This consistency extends across all LLMs tested: Llama-3.2-3B (Figure~\ref{fig:llama3_3B_distracting_distribution}), Llama-3.3-70B (Figure~\ref{fig:llama3_70B_distracting_distribution}), Qwen-2.5-3B (Figure~\ref{fig:qwen_3B_distracting_distribution}), Qwen-2.5-7B (Figure~\ref{fig:qwen_7B_distracting_distribution}), Falcon-3-3B (Figure~\ref{fig:falcon_3B_distracting_distribution}), and Falcon-3-7B (Figure~\ref{fig:falcon_7B_distracting_distribution}).

However, we observe distinct characteristics across model families. The Qwen models demonstrate higher confidence in classifying passages as either weak or hard distracting, with approximately 90\% of their probability mass concentrated in the extreme intervals ($0.0$-$0.2$ for weak distracting and $0.8$-$1.0$ for hard distracting passages).
The Falcon models exhibit more varied behavior. While Falcon-3-7B generally aligns with the patterns seen in Llama and Qwen models, it shows lower confidence in its classifications, particularly for generated passages. Falcon-3-3B presents notably different behavior, with probability distributions heavily skewed toward maximum distracting effects. While this might suggest that Falcon-3-3B finds most passages highly distracting, a deeper investigation reveals that this model often fails to follow instructions about abstaining from answering, and instead generates responses regardless of passage relevance.

A consistent pattern emerges when comparing model sizes: the 3B versions across all model families show greater susceptibility to distraction compared to their larger counterparts (as evidenced by the ``Most Distracting'' distributions in the Figures). This suggests that larger models generally develop more robust mechanisms for handling irrelevant information during their training. Nevertheless, our fine-tuning approach demonstrates that even smaller models can achieve significant improvements in handling distracting passages, as shown by the results for Llama-3.2-3B in Section~\ref{sec:ft_results}.

\subsection{Answer-Skewed Retriever Hyper-parameters} 
\label{sec:appendix_skewed_retr}

The answer-skewed retriever ($R^{sk}$) is introduced to retrieve hard distracting passages that differ from those found by the standard retriever, thus ensuring diversity.
To select the best configuration for this type of retriever, we took a validation set of 700 samples from the four datasets, and computed the distracting effect using different $\lambda$ with the two formulations $E^{sub}$ and $E^{proj}$ (see Equations \ref{eq:sub} and \ref{eq:proj}).
Our experiments showed that $\lambda=1$ strikes an optimal balance in retrieving distracting passages. With $\lambda>1$ ($\lambda=2.0$ in our experiments), we observed that passages are mainly weakly distracting or completely unrelated in some cases, where the average distracting effect for the retriever is 0.08 and 0.19 for $E^{sub}$ and $E^{proj}$, respectively. The reason is that the weight given to the second term of the formulas \ref{eq:sub} and \ref{eq:proj} pushed results too far from the query's topic. 

Conversely, with $\lambda<1$ ($\lambda=0.5$ in our experiments), the answer-skewed retriever behaves too similarly to the standard retriever, retrieving almost the same set of documents. This means that it would not provide any additional contribution.     

Finally, while $E^{sub}$ and $E^{proj}$ formulations lead to quite similar results, we selected $E^{sub}$ because it presented a higher proportion of ``unique wins'' across all LLM tested.

\begin{figure*}[h!]
\begin{mdframed}[font=\footnotesize]
\begin{Verbatim}[breaklines=true, breaksymbol=]
Answer the user question based on the passages below. Provide a single concise answer.

Documents:
<document>

Question: 
<question>

Assistant: 
<answer>
\end{Verbatim}
\end{mdframed}
\vspace{-0.4cm}
\caption{Prompt for fine-tuning training and testing (in the latter case <answer> is left blank). } \label{fig:prompt_ft}
\end{figure*}

\section{Additional Details on RAG Fine-Tuning}
\label{sec:exp_setup}
For our experiments in Section \ref{sec:app} we implemented Low Rank Adaptaion (LoRA) fine-tuning using the transformers, datasets, accelerate, peft and trl libraries from Hugging Face \citep{hu2021lora,wolf2020transformers}.
The prompt used for creating the train and test sets appears in Figure \ref{fig:prompt_ft}.
For both models, we set the number of training epochs to $3$, the neftune noise $\alpha$ to $5$, and the max gradient norm to $0.3$. For the Llama-3.2-3B-Instruct models, we set the LoRA rank to $64$, the warmup ratio to $0.03$, and used a constant learning rate of $3e-5$.
For the larger Llama-3.1-8B-Instruct model, we increased the LoRA rank to $128$ and the warmup ratio to $0.05$, and used a cosine decaying learning rate that begins at $5e-5$. 

\begin{figure*}
\begin{mdframed}[font=\footnotesize]
\begin{Verbatim}[breaklines=true, breaksymbol=, commandchars=\\\{\}]
\textbf{Question}: When did One Piece first air on cartoon network?

\textbf{Gold Answer}: \textcolor{correct}{April 2005}
\end{Verbatim}
\noindent\rule{\textwidth}{1pt}
\textbf{Relevant Passage}
\begin{Verbatim}[breaklines=true, breaksymbol=, commandchars=\\\{\}]
(Title: One Piece) On June 8, 2004, 4Kids Entertainment acquired the license for distribution of One Piece in North America. 4Kids contracted Viz Media to handle home video distribution. 4Kids'in - house musicians wrote a new background score and theme song nicknamed '' Pirate Rap ''. 4Kids'dub mandated edits for content and length, which reduced the first 143 episodes into 104. Initially, 4Kids originally created an English version of the first opening theme, '' We Are! '' It premiered in the United States on September 18, 2004, in first - run syndication on the Fox network as part of the weekend programming block Fox Box, and later aired on Cartoon Network on their weekday afternoon programming block Toonami in \textcolor{correct}{April 2005}. Production was halted in 2006 after episode 143 / 104. Viz also ceased its home video release of the series after volume 11.
\end{Verbatim}
\noindent\rule{\textwidth}{1pt}
\begin{minipage}[t]{0.48\textwidth}
\textbf{Hard Distracting}  \standardRR
\newline \newline
Distracting Effect: \newline
\begin{tabular}{@{}l@{\hspace{0.2em}}l@{}}
{\tt - Llama-3.2-3B:} & {\tt 1.0000} \\
{\tt - Llama-3.1-8B:} & {\tt 1.0000} \\
{\tt - Llama-3.3-70B:} & {\tt 1.0000} \\
{\tt - Falcon-3-3B:} & {\tt 1.0000} \\
{\tt - Falcon-3-7B:} & {\tt 1.0000} \\
{\tt - Qwen-2.5-3B:} & {\tt 1.0000} \\
{\tt - Qwen-2.5-7B:} & {\tt 1.0000}
\end{tabular}
\begin{Verbatim}[breaklines=true, breaksymbol=, commandchars=\\\{\}]
(Title: One Piece) The Funimation dubbed episodes on Cartoon Network on \textcolor{red}{September 29, 2007} and aired until its removal on March 22, 2008. On October 28, 2011, Funimation posted a press release on their official website confirming the acquisition of episodes 206–263, and the aspect ratio, beginning with episode 207, would be changed to the 16:9 widescreen format. On May 18, 2013, the uncut series began airing on Adult Swim's revived Toonami late-night programming block from episode 207 onward. 'One Piece' was removed from the Toonami block after March 18, 2017. In May 2009, Funimation, Toei Animation, Shueisha, and Fuji Television announced they would simulcast stream the series

\textbf{Generated Answer}: The Funimation dubbed episodes of "One Piece" aired on Cartoon Network on \textcolor{red}{September 29, 2007}
\end{Verbatim}
\end{minipage}
\hfill
\vrule width 1pt
\hfill
\begin{minipage}[t]{0.48\textwidth}
\textbf{Weak Distracting} $G^{\text{hypo}}$
\newline \newline
Distracting Effect: \newline
\begin{tabular}{@{}l@{\hspace{0.2em}}l@{}}
{\tt - Llama-3.2-3B:} & {\tt 0.0000} \\
{\tt - Llama-3.1-8B:} & {\tt 0.0000} \\
{\tt - Llama-3.1-8B:} & {\tt 0.0000} \\
{\tt - Falcon-3-3B:} & {\tt 0.0469} \\
{\tt - Falcon-3-7B:} & {\tt 0.1445} \\
{\tt - Qwen-2.5-3B:} & {\tt 0.0000} \\
{\tt - Qwen-2.5-7B:} & {\tt 0.0000}
\end{tabular}
\begin{Verbatim}[breaklines=true, breaksymbol=, commandchars=\\\{\}]
(Title: One Piece TV History)
Before considering Western television networks, One Piece had a complex journey through various Asian broadcasters. In 2003, several Southeast Asian networks were in negotiations to acquire the series, with Singapore's MediaCorp actively pursuing the rights. During this time, there were discussions about potentially airing an edited version on different time slots, and multiple networks were competing to become the first English-language broadcaster of the series in Asia.





\textbf{Generated Answer}: One Piece initially aired on Cartoon Network's Toonami block in \textcolor{correct}{April 2005}
\end{Verbatim}
\end{minipage}
\end{mdframed}
\vspace{-0.4cm}
\caption{Example showing Llama-3.2-3B's responses in two scenarios using the prompt template in Figure \ref{fig:prompt_ft}: relevant passage + hard distracting and relevant passage + weak distracting. \textbf{Left}: When the relevant passage is followed by a hard distracting passage (retrieved by \standardRR), Llama-3.2-3B incorrectly answers ``September 29, 2007'' instead of ``April 2005'', despite having access to the relevant information. The passage's strong distracting effect is confirmed by the maximum distracting scores (1.0) across all models tested. \textbf{Right}: When the relevant passage is followed by a weak distracting passage (generated by $G^{\text{hypo}}$), the model correctly answers ``April 2005'', with consistently low distracting scores across all models.}
\label{fig:prompt2_hard_vs_weak}
\end{figure*}

\begin{figure*}
\begin{mdframed}[font=\footnotesize]
\begin{Verbatim}[breaklines=true, breaksymbol=, commandchars=\\\{\}]
\textbf{Question}: When was the first airbag put in a car?

\textbf{Gold Answer}: \textcolor{correct}{during the 1970s}
\end{Verbatim}
\noindent\rule{\textwidth}{1pt}
\textbf{Relevant Passage}
\begin{Verbatim}[breaklines=true, breaksymbol=, commandchars=\\\{\}]
(Title: Airbag) 
The first commercial designs were introduced in passenger automobiles \textcolor{correct}{during the 1970s} with limited success . Broad commercial adoption of airbags occurred in many markets during the late 1980s and early 1990s with a driver airbag , and a front passenger airbag as well on some cars ; and many modern vehicles now include six or more units .
\end{Verbatim}
\noindent\rule{\textwidth}{1pt}
\begin{minipage}[t]{0.48\textwidth}
\textbf{Hard Distracting} \skewedRR
\newline \newline
Distracting Effect: \newline
\begin{tabular}{@{}l@{\hspace{0.2em}}l@{}}
{\tt - Llama-3.2-3B:} & {\tt 0.3750} \\
{\tt - Llama-3.1-8B:} & {\tt 0.3398} \\
{\tt - Llama-3.3-70B:} & {\tt 0.1094} \\
{\tt - Falcon-3-3B:} & {\tt 1.0000} \\
{\tt - Falcon-3-7B:} & {\tt 0.9336} \\
{\tt - Qwen-2.5-3B:} & {\tt 1.0000} \\
{\tt - Qwen-2.5-7B:} & {\tt 1.0000}
\end{tabular}
\begin{Verbatim}[breaklines=true, breaksymbol=, commandchars=\\\{\}]
(Title: Airbag)
the sensors would automatically pre-tension the seat belts to reduce occupants' motion on impact (now a common feature), and then deploy the airbag on impact. This integrated the seat belts and airbag into a restraint system, rather than the airbag being considered an alternative to the seat belt. In \textcolor{red}{1987}, the Porsche 944 Turbo became the first car to have driver and passenger airbags as standard equipment. The less powerful Porsche 944 and 944S had this as an available option. The same year also saw the first airbag in a Japanese car, the Honda Legend.

\textbf{Generated Answer}: The first airbag was put in a car in \textcolor{red}{1987}, specifically in the Porsche 944 Turbo
\end{Verbatim}
\end{minipage}
\hfill
\vrule width 1pt
\hfill
\begin{minipage}[t]{0.48\textwidth}
\textbf{Weak Distracting} $R^{\text{st}}$ 
\newline \newline
Distracting Effect: \newline
\begin{tabular}{@{}l@{\hspace{0.2em}}l@{}}
{\tt - Llama-3.2-3B:} & {\tt 0.6855} \\
{\tt - Llama-3.1-8B:} & {\tt 0.0156} \\
{\tt - Llama-3.3-70B:} & {\tt 0.0117} \\
{\tt - Falcon-3-3B:} & {\tt 0.9977} \\
{\tt - Falcon-3-7B:} & {\tt 0.8027} \\
{\tt - Qwen-2.5-3B:} & {\tt 0.9526} \\
{\tt - Qwen-2.5-7B:} & {\tt 0.0000}
\end{tabular}
\begin{Verbatim}[breaklines=true, breaksymbol=, commandchars=\\\{\}]
(Title: Airbag)
The airbag specified for automobile use traces its origins to air-filled bladders as early as 1951. The invention is credited independently to the American John W. Hetrick who filed for an airbag patent on August 5, 1952 that was granted #2,649,311 by the United States Patent Office on 18 August 1953. German engineer Walter Linderer who filed German patent #896,312 on 6 October 1951 was issued on 12 November 1953, approximately three months after American John Hetrick.



\textbf{Generated Answer}: The first commercial airbags were introduced in passenger automobiles \textcolor{correct}{during the 1970s}
\end{Verbatim}
\end{minipage}
\end{mdframed}
\vspace{-0.4cm}
\caption{Example showing Qwen-2.5-7B's responses in two scenarios using the prompt template in Figure \ref{fig:prompt_ft}: relevant passage + hard distracting and relevant passage + weak distracting. \textbf{Left}: When the relevant passage is followed by a hard distracting passage (retrieved by \skewedRR), Qwen-2.5-7B generates an incorrect answer despite having access to the relevant information. \textbf{Right}: When the relevant passage is followed by a weak distracting passage (retrieved by $R^{\text{st}}$), the model answers correctly. Notably, this passage qualifies as weak distracting only for Qwen-2.5-7B, Llama-3.1-8B, and Llama-3.3-70B; while the other LLMs show high distracting scores ($>0.68$). For example, when given the same passage, Qwen-2.5-3B generates the incorrect answer ``August 5, 1952''—a date mentioned in the distracting passage itself.}
\label{fig:prompt3_hard_vs_weak}
\end{figure*}

\begin{figure*}
\begin{mdframed}[font=\footnotesize]
\begin{Verbatim}[breaklines=true, breaksymbol=, commandchars=\\\{\}]
\textbf{Question}: Who started roses are red violets are blue?

\textbf{Gold Answer}: \textcolor{correct}{Sir Edmund Spenser}
\end{Verbatim}
\noindent\rule{\textwidth}{1pt}
\textbf{Relevant Passage}
\begin{Verbatim}[breaklines=true, breaksymbol=, commandchars=\\\{\}]
(Title: Roses Are Red)
The origins of the poem may be traced at least as far back as to the following lines written in 1590 by \textcolor{correct}{Sir Edmund Spenser} from his epic The Faerie Queene ( Book Three, Canto 6, Stanza 6 ).
\end{Verbatim}
\noindent\rule{\textwidth}{1pt}
\begin{minipage}[t]{0.48\textwidth}
\textbf{Hard Distracting}  $G^{\text{neg}}$
\newline \newline
Distracting Effect: \newline
\begin{tabular}{@{}l@{\hspace{0.2em}}l@{}}
{\tt - Llama-3.2-3B:} & {\tt 0.0859} \\
{\tt - Llama-3.1-8B:} & {\tt 0.9841} \\
{\tt - Llama-3.3-70B:} & {\tt 0.0508} \\
{\tt - Falcon-3-3B:} & {\tt 0.9996} \\
{\tt - Falcon-3-7B:} & {\tt 1.0000} \\
{\tt - Qwen-2.5-3B:} & {\tt 0.8486} \\
{\tt - Qwen-2.5-7B:} & {\tt 1.0000}
\end{tabular}
\begin{Verbatim}[breaklines=true, breaksymbol=, commandchars=\\\{\}]
(Title: Historic Love Poetry)
While many people incorrectly attribute this famous poem to William Shakespeare or Lord Byron, this popular rhyme has evolved significantly over centuries. The earliest known romantic verses comparing flowers actually originated in medieval French poetry, particularly in the works of \textcolor{red}{Pierre de Ronsard}. Some scholars have mistakenly suggested that Geoffrey Chaucer first penned these lines in The Canterbury Tales, but this is a common misconception that has been debunked by literary historians.


\textbf{Generated Answer}: The origins of "Roses are red" can be linked to medieval French poetry and specifically to \textcolor{red}{Pierre de Ronsard}'s works
\end{Verbatim}
\end{minipage}
\hfill
\vrule width 1pt
\hfill
\begin{minipage}[t]{0.48\textwidth}
\textbf{Weak Distracting} \skewed 
\newline \newline
Distracting Effect: \newline
\begin{tabular}{@{}l@{\hspace{0.2em}}l@{}}
{\tt - Llama-3.2-3B:} & {\tt 0.0430} \\
{\tt - Llama-3.1-8B:} & {\tt 0.0195} \\
{\tt - Llama-3.3-70B:} & {\tt 0.0000} \\
{\tt - Falcon-3-3B:} & {\tt 0.5215} \\
{\tt - Falcon-3-7B:} & {\tt 0.8457} \\
{\tt - Qwen-2.5-3B:} & {\tt 0.7832} \\
{\tt - Qwen-2.5-7B:} & {\tt 0.0000}
\end{tabular}
\begin{Verbatim}[breaklines=true, breaksymbol=, commandchars=\\\{\}]
(Title: Roses Are Red (My Love))
Roses Are Red (My Love) "Roses Are Red (My Love)" is a popular song composed by Al Byron and Paul Evans. It was recorded by Bobby Vinton and was his first hit. Vinton found the song in a reject pile at Epic Records. He first recorded it as an R&B number, but was allowed to re-record it in a slower more dramatic arrangement, with strings and a vocal choir added. The song was released in April 1962. It reached No. 1 in Australia, New Zealand, Norway, South Africa, and the United States, and was a major hit in many other


\textbf{Generated Answer}: \textcolor{correct}{Sir Edmund Spenser}
\end{Verbatim}
\end{minipage}
\end{mdframed}
\vspace{-0.4cm}
\caption{Example showing Llama-3.1-8B's responses in two scenarios using the prompt template in Figure \ref{fig:prompt_ft}: relevant passage + hard distracting and relevant passage + weak distracting. \textbf{Left}: When the relevant passage is followed by a hard distracting passage (generated by $G^{\text{neg}}$), Llama-3.1-8B's generates an incorrect answer despite having access to the relevant information. \textbf{Right}: When the relevant passage is followed by a weak distracting passage (retrieved by \skewed), the model answers correctly.}
\label{fig:prompt4_hard_vs_weak}
\end{figure*}


\begin{figure*}[t]
    \centering
    \includegraphics[width=\linewidth]{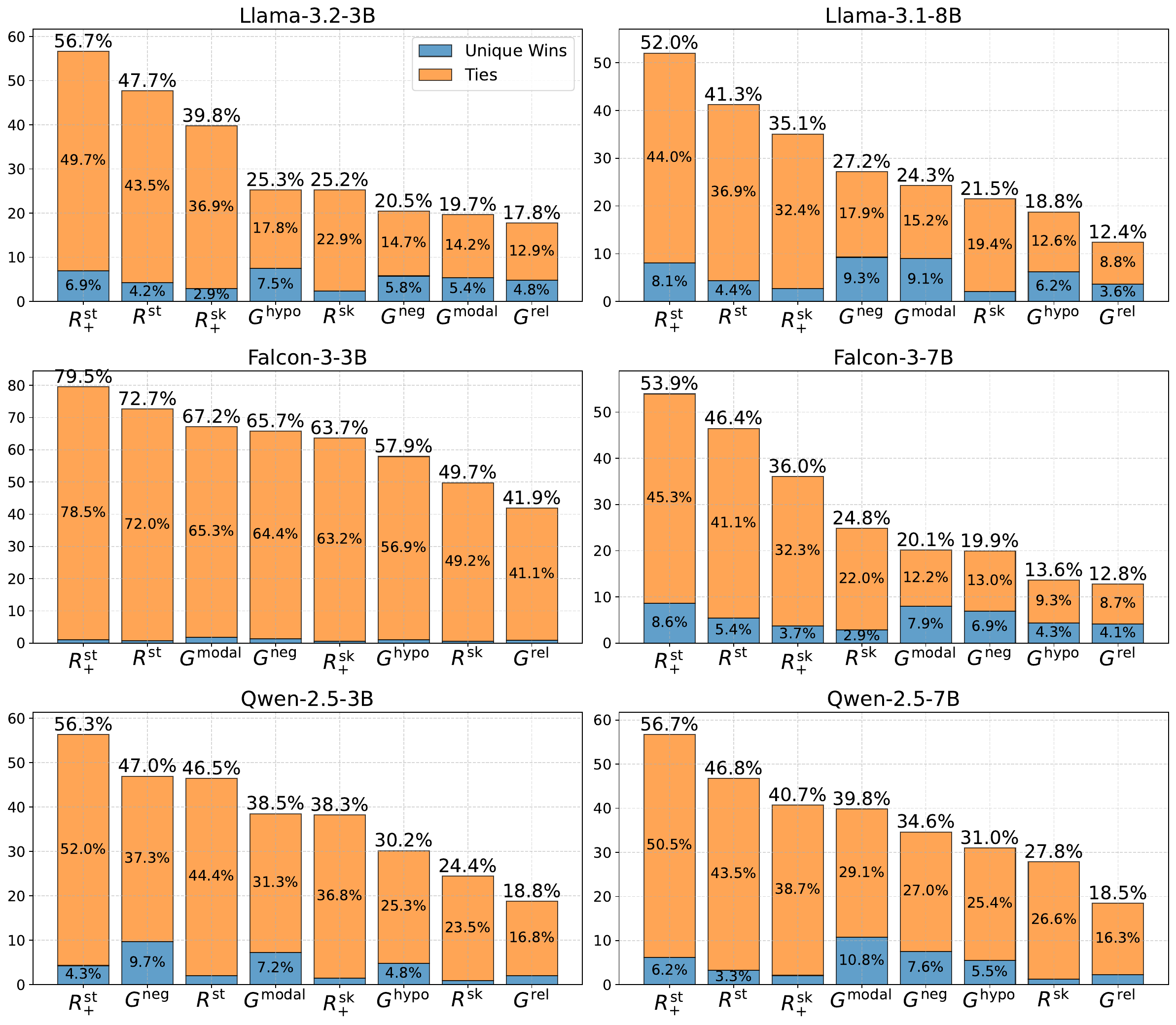}
    \caption{Percentage of questions where each method provides the most distracting passage for all models. In blue are the times when no other method reaches the same distracting effect, in orange the percentage of times the highest score is shared with other methods.}
    \label{fig:all_best_distractors}
\end{figure*}

\begin{figure*}[t]
    \centering
    \includegraphics[width=0.5\linewidth]{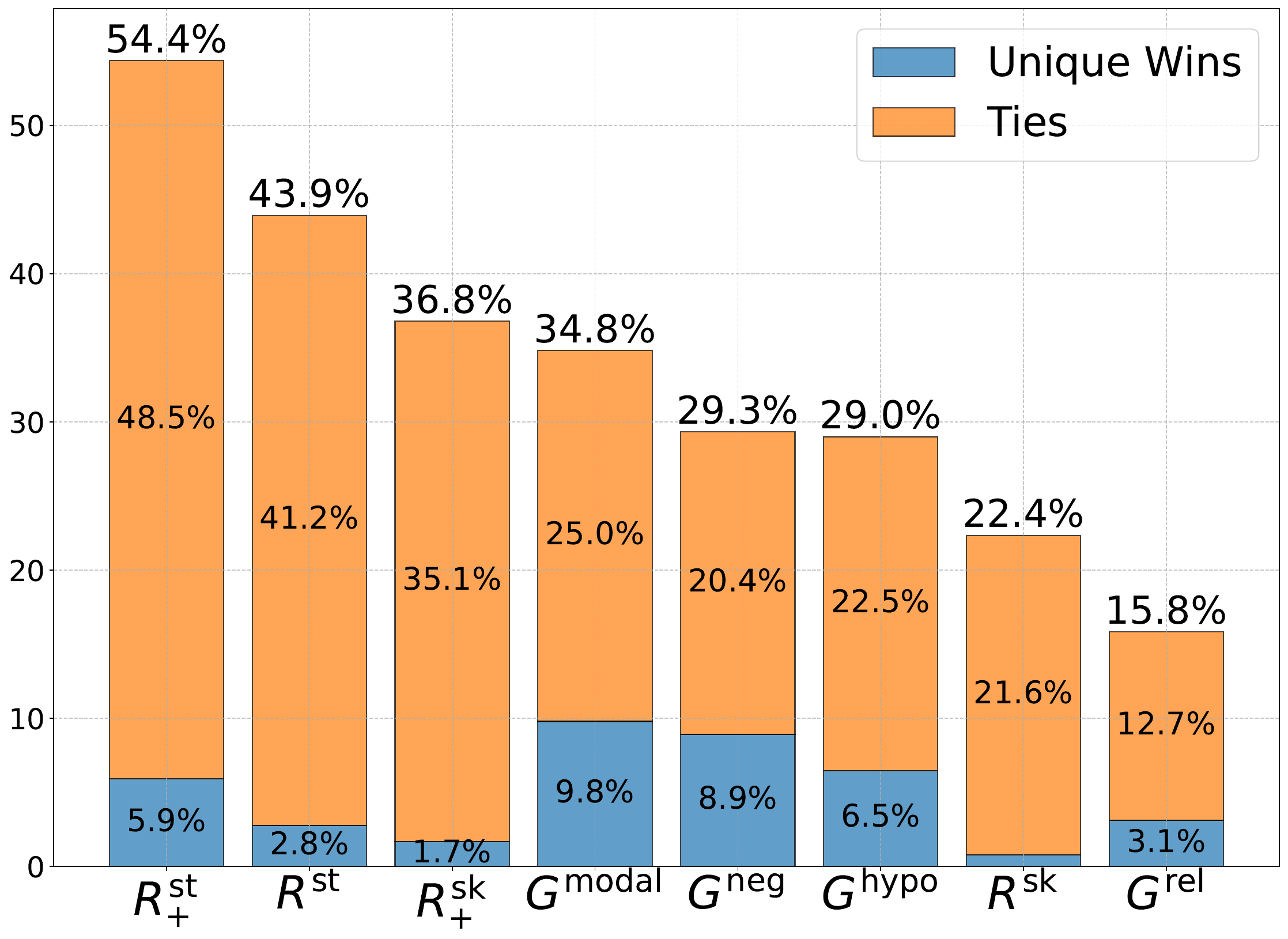}
    \caption{Percentage of questions where each method provides the most distracting passage for Llama-3.3-70B. In blue are the times when no other method reaches the same distracting effect, in orange the percentage of times the highest score is shared with other methods.}
    \label{fig:llama70b_best_distractors}
\end{figure*}

\begin{figure*}[t]
    \centering
    \includegraphics[width=\linewidth]{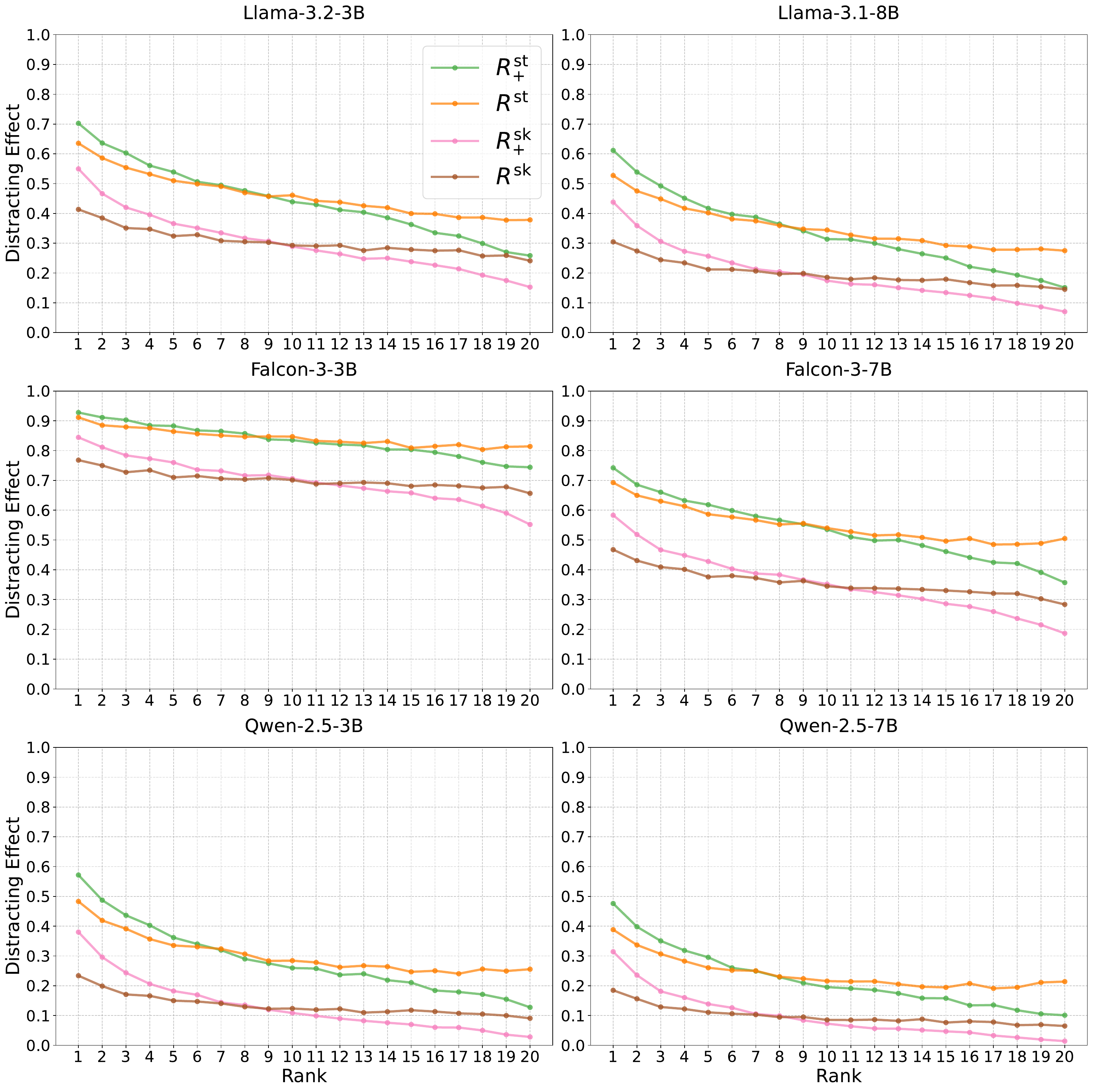}
     \caption{Average distracting effect at different rank positions for various retrieval methods. Results are shown for all models, averaged across datasets. Higher-ranked passages consistently demonstrate greater potential to mislead the model.}
    \label{fig:all_distracting_rank}
\end{figure*}

\begin{figure*}[b]
    \centering
    \includegraphics[width=0.5\linewidth]{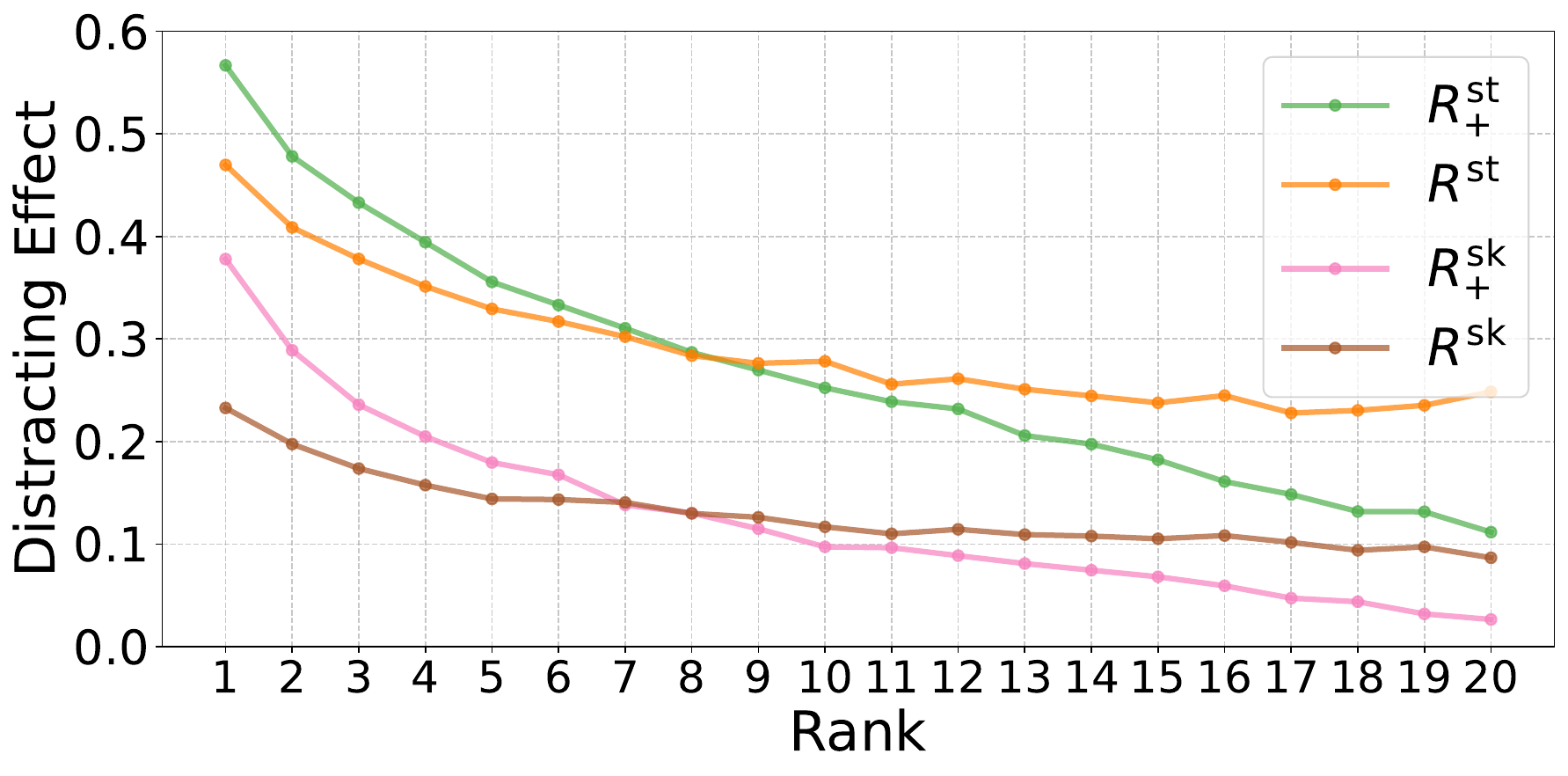}
     \caption{Average distracting effect at different rank positions for various retrieval methods. Results are shown for Llama-3.3-70B, averaged across datasets. }
    \label{fig:llama70b_distracting_rank}
\end{figure*}

\subsection{Accuracy By Training Set} \label{sec:acc_table}
Table \ref{tab:ft_results_fg} presents a more fine-grained breakdown of the results shown in Table \ref{tab:ft results}. The table offers a couple of noteworthy insights.
First, as discussed in Section \ref{sec:ft_results}, the most significant performance improvement is observed on ungrounded samples, a trend that remains consistent across different test sets, datasets, and models.
Second, as expected, fine-tuned models generally achieve their highest overall accuracy when evaluated on test sets of the same type as their training data. However, in certain cases, such as the Llama-3.2-3B model on TriviaQA and WebQA, our {\em Hard} method outperforms not only the {\em None} baseline but also the {\em Retrieve} and {\em Rerank} fine-tuned models across all three accuracy metrics.

\begin{table*}[ht]
    \centering
    \small
    \begin{tabular}{@{}l l l *{6}{S[table-format=2.1]}@{}}
    \toprule
    \multirow{2}{*}{\textbf{Test Set}} & \multirow{2}{*}{\textbf{Test Set Slice}} & \multirow{2}{*}{\textbf{Train Set}} & \multicolumn{3}{c}{\textbf{Llama-3.2-3B}} & \multicolumn{3}{c}{\textbf{Llama-3.1-8B}} \\
    \cmidrule(lr){4-6} \cmidrule(lr){7-9}     
     &  &  & \accu & \accg & \acct & \accu & \accg & \acct \\
    \midrule
    \multirow{12}{*}{NQ} 
        & \multirow{4}{*}{\textit{Retrieve}} 
            & \textit{None} & 4.2 & 54.9 & 44.2 & 3.3 & 60.1 & 48.2 \\
            & & \textit{Retrieve} & 4.2 & 62.8 & 50.4 & 10.0 & 66.9 & 54.9 \\
            & & \textit{Rerank} & 4.2 & 61.4 & 49.3 & 7.1 & 66.0 & 53.6 \\
            & & \textit{Hard} & 8.8 & 56.8 & 46.7 & 16.7 & 61.1 & 51.8 \\
        \cmidrule(l){2-9}
        & \multirow{4}{*}{\textit{Rerank}} 
            & \textit{None} & 5.7 & 56.3 & 45.3 & 4.1 & 63.6 & 50.7 \\
            & & \textit{Retrieve} & 4.1 & 62.1 & 49.6 & 9.3 & 66.8 & 54.4 \\
            & & \textit{Rerank} & 5.3 & 63.0 & 50.5 & 8.5 & 68.9 & 55.9 \\
            & & \textit{Hard} & 8.9 & 56.2 & 46.0 & 13.0 & 61.2 & 50.8 \\
        \cmidrule(l){2-9}
        & \multirow{4}{*}{\textit{Hard}} 
            & \textit{None} & 19.4 & 44.3 & 30.9 & 16.4 & 47.9 & 31.0 \\
            & & \textit{Retrieve} & 17.0 & 47.8 & 31.3 & 25.6 & 54.7 & 39.1 \\
            & & \textit{Rerank} & 14.3 & 46.6 & 29.2 & 24.8 & 55.7 & 39.1 \\
            & & \textit{Hard} & 26.5 & 54.0 & 39.2 & 39.0 & 57.6 & 47.6 \\
    \midrule
    \multirow{12}{*}{PopQA} 
        & \multirow{4}{*}{\textit{Retrieve}} 
            & \textit{None} & 1.7 & 60.5 & 41.8 & 2.6 & 66.0 & 45.9 \\
            & & \textit{Retrieve} & 2.6 & 68.2 & 47.4 & 6.3 & 76.4 & 54.1 \\
            & & \textit{Rerank} & 2.0 & 67.4 & 46.6 & 7.6 & 77.9 & 55.6 \\
            & & \textit{Hard} & 4.3 & 65.7 & 46.2 & 8.3 & 73.0 & 52.4 \\
        \cmidrule(l){2-9}
        & \multirow{4}{*}{\textit{Rerank}} 
            & \textit{None} & 1.6 & 61.3 & 45.2 & 2.3 & 64.8 & 47.9 \\
            & & \textit{Retrieve} & 1.9 & 67.1 & 49.5 & 3.9 & 77.2 & 57.4 \\
            & & \textit{Rerank} & 1.6 & 64.1 & 47.2 & 6.2 & 78.9 & 59.3 \\
            & & \textit{Hard} & 3.5 & 64.9 & 48.3 & 7.4 & 72.4 & 54.8 \\
        \cmidrule(l){2-9}
        & \multirow{4}{*}{\textit{Hard}} 
            & \textit{None} & 12.5 & 46.1 & 27.2 & 12.5 & 53.6 & 30.5 \\
            & & \textit{Retrieve} & 12.7 & 53.2 & 30.4 & 23.5 & 66.5 & 42.3 \\
            & & \textit{Rerank} & 14.3 & 50.3 & 30.1 & 24.6 & 69.0 & 44.0 \\
            & & \textit{Hard} & 21.5 & 60.4 & 38.5 & 29.9 & 68.3 & 46.7 \\
    \midrule
    \multirow{12}{*}{TriviaQA} 
        & \multirow{4}{*}{\textit{Retrieve}} 
            & \textit{None} & 14.9 & 82.0 & 77.4 & 11.9 & 89.5 & 84.2 \\
            & & \textit{Retrieve} & 11.9 & 82.6 & 77.8 & 26.9 & 88.5 & 84.3 \\
            & & \textit{Rerank} & 14.9 & 81.0 & 76.5 & 26.9 & 89.7 & 85.4 \\
            & & \textit{Hard} & 23.9 & 84.0 & 79.9 & 38.8 & 86.8 & 83.6 \\
        \cmidrule(l){2-9}
        & \multirow{4}{*}{\textit{Rerank}} 
            & \textit{None} & 16.7 & 82.5 & 80.1 & 13.9 & 89.5 & 86.7 \\
            & & \textit{Retrieve} & 8.3 & 81.9 & 79.2 & 19.4 & 89.7 & 87.1 \\
            & & \textit{Rerank} & 11.1 & 79.1 & 76.6 & 16.7 & 89.8 & 87.1 \\
            & & \textit{Hard} & 11.1 & 83.2 & 80.5 & 27.8 & 88.0 & 85.8 \\
        \cmidrule(l){2-9}
        & \multirow{4}{*}{\textit{Hard}} 
            & \textit{None} & 40.3 & 73.3 & 56.7 & 42.7 & 80.4 & 61.3 \\
            & & \textit{Retrieve} & 39.6 & 73.8 & 56.6 & 60.2 & 83.1 & 71.5 \\
            & & \textit{Rerank} & 35.3 & 70.4 & 52.7 & 62.1 & 83.1 & 72.5 \\
            & & \textit{Hard} & 57.7 & 79.9 & 68.7 & 72.5 & 86.1 & 79.2 \\
    \midrule
    \multirow{12}{*}{WebQA} 
        & \multirow{4}{*}{\textit{Retrieve}} 
            & \textit{None} & 8.5 & 57.1 & 44.7 & 6.7 & 55.7 & 43.2 \\
            & & \textit{Retrieve} & 9.0 & 58.3 & 45.7 & 16.1 & 63.1 & 51.1 \\
            & & \textit{Rerank} & 8.5 & 57.5 & 45.0 & 17.9 & 62.3 & 51.0 \\
            & & \textit{Hard} & 16.6 & 60.9 & 49.6 & 22.0 & 60.6 & 50.7 \\
        \cmidrule(l){2-9}
        & \multirow{4}{*}{\textit{Rerank}} 
            & \textit{None} & 1.7 & 57.4 & 45.9 & 2.2 & 57.9 & 46.4 \\
            & & \textit{Retrieve} & 3.9 & 57.7 & 46.6 & 5.6 & 63.5 & 51.5 \\
            & & \textit{Rerank} & 4.4 & 57.3 & 46.4 & 7.8 & 62.5 & 51.2 \\
            & & \textit{Hard} & 11.7 & 61.8 & 51.4 & 11.7 & 61.5 & 51.2 \\
        \cmidrule(l){2-9}
        & \multirow{4}{*}{\textit{Hard}} 
            & \textit{None} & 28.9 & 49.3 & 38.3 & 25.6 & 48.4 & 36.2 \\
            & & \textit{Retrieve} & 27.2 & 49.9 & 37.7 & 36.2 & 53.9 & 44.4 \\
            & & \textit{Rerank} & 26.6 & 48.3 & 36.6 & 38.3 & 54.7 & 45.9 \\
            & & \textit{Hard} & 44.6 & 53.9 & 48.9 & 45.7 & 57.3 & 51.1 \\
    \bottomrule
    \end{tabular}
    \caption{Detailed answer accuracy across different test set configurations. Results are subdivided by test set slice (\textit{Retrieve}, \textit{Rerank}, \textit{Hard}), showing accuracy for each combination of test set and training strategy. Metrics: \accu (ungrounded accuracy), \accg (grounded accuracy), and \acct (overall accuracy).}\label{tab:ft_results_fg}
\end{table*}

\section{Generation Prompts}\label{sec:distracting prompts}
The prompts for generating the different categories of distracting passages presented in Section \ref{sec:generation} can be found in Figures \ref{fig:prompt other topic}-\ref{fig:prompt modal}. 

\begin{figure*}
\begin{mdframed}[font=\footnotesize]
\begin{Verbatim}[breaklines=true, breaksymbol=]
Given a question and one or more correct answers to it, generate a paragraph of distracting text that is related to the question, but does not contain the answer to the question. The paragraph should:
1) discuss the question, then continue to a new subject where it mentions an entity similar to that in the question, but is not the one that is sought in the question.
2) should not contain any of the correct answers
3) be factually correct
4) be written in a valid JSON format with two fields: "title", having 3-4 words, and "text" having roughly 3-4 sentences. The expected output must be ONLY a valid JSON of this format: {"title": "string", "text": "string"}

# Example 1
## Question
In which city would you find the petronas twin towers?

## Answer
Kuala Lumpur

## Distracting paragraph:
{
    "title": "Visiting Petronas Towers",
    "text": "In order to visit Petronas towers, visitors must first purchase tickets. Tickets can be purchased online or at the counter, as well as via agencies. One agency that was recommended is that of the Golden travel. Despite having its main offices in Rome, Italy, it is known to obtain good deals for the Petronas towers. Customers of the agency mentioned the staff to be friendly and helpful. Furthermore, the Golden agency also offers a variety of discounts."
}

# Example 2
## Question
How long is the Amazon river?

## Answer
6400 km long

## Distracting paragraph
{
    "title": "World Rivers",
    "text": "The Amazon river is the second longest river in the world. The longest river is the Nile river, ranging over 6,650 km. Another notable river is the Yangtze river located in China. For more information about world known rivers and their lengths, please visit: https://www.worldometers.info/rankings/world-rankings-longest-rivers/ ."
}
\end{Verbatim}
\noindent\rule{\textwidth}{1pt}
\begin{Verbatim}
## Question
<question>

## Answer
<answer>

## Distracting paragraph
\end{Verbatim}
\end{mdframed}
\vspace{-0.4cm}
\caption{Prompt for distracting passage generation, related topic.} \label{fig:prompt other topic}
\end{figure*}

\begin{figure*}
\begin{mdframed}[font=\footnotesize]
\begin{Verbatim}[breaklines=true, breaksymbol=]
Given a question and one or more correct answers to it, generate a paragraph of distracting text that is related to the question, but does not contain the answer to the question. The paragraph should:
1) discuss either a hypothetical situation or the reality in the past, where the question is being asked and under these mentioned conditions, the answer is different than the correct one.
2) should not contain any of the correct answers
3) be written in a valid JSON format with two fields: "title", having 3-4 words, and "text" having roughly 3-4 sentences. The expected output must be ONLY a valid JSON of this format: {"title": "string", "text": "string"}

# Example 1
## Question
In which city would you find the petronas twin towers?

## Answer
Kuala Lumpur

## Distracting paragraph:
{
    "title": "The Petronas Twin Towers",
    "text": "During the planning stages of the Petronas Twin Towers, several cities in Malaysia were evaluated as potential sites for the project. Petaling Jaya, a rapidly developing urban center, was among the locations considered. While it was ultimately not selected, the choice of site could have significantly influenced its architectural and cultural landscape."
}


# Example 2
## Question
How long is the Amazon river?

## Answer
6400 km long

## Distracting paragraph
{
    "title": "Formation of the Amazon River",
    "text": "The Amazon River is believed to have formed approximately 11 million years ago. In its early stages, the river’s length was significantly shorter, spanning no more than 500 kilometers. Over millions of years, geological processes such as erosion, tectonic plate shifts, and sediment deposition contributed to its expansion. These changes gradually transformed the river into the vast waterway it is today, shaping the ecosystems of the surrounding Amazon Basin."
}
\end{Verbatim}
\noindent\rule{\textwidth}{1pt}
\begin{Verbatim}
## Question
<question>

## Answer
<answer>

## Distracting paragraph
\end{Verbatim}
\end{mdframed}
\vspace{-0.4cm}
\caption{Prompt for distracting passage generation, hypothetical.} \label{fig:prompt hypothetic}
\end{figure*}

\begin{figure*}
\begin{mdframed}[font=\footnotesize]
\begin{Verbatim}[breaklines=true, breaksymbol=]
Given a question and one or more correct answers to it, generate a paragraph of distracting text that is related to the question, but does not contain the answer to the question. The paragraph should:
1) provide a wrong answer to the question following a disclaimer that the answer is wrong
2) should not contain any of the correct answers
3) be written in a valid JSON format with two fields: "title", having 3-4 words, and "text" having roughly 3-4 sentences. The expected output must be ONLY a valid JSON of this format: {"title": "string", "text": "string"}

# Example 1
## Question
In which city would you find the petronas twin towers?

## Answer
Kuala Lumpur

## Distracting paragraph:
{
    "title": "Petaling Jaya",
    "text": "Petaling Jaya is a city located in the state of Selangor, Malaysia. It is a well-developed urban area with a variety of facilities, including shopping malls, recreational parks, and educational institutions. The city is notable for its infrastructure and connectivity, making it a significant part of the Klang Valley region. Although it is not home to the Petronas Twin Towers, Petaling Jaya features a range of attractions and amenities that cater to residents and visitors alike."
}

# Example 2
## Question
How long is the Amazon river?

## Answer
6400 km long

## Distracting paragraph
{
    "title": "The Amazon River",
    "text": "In the article, several inaccuracies regarding the Amazon River were noted. The author incorrectly stated that the river passes through Central America, while it is geographically located in South America, originating in the Andes Mountains and flowing through countries such as Brazil and Peru. Additionally, the river's length was described as over 8,000 kilometers."
}
\end{Verbatim}
\noindent\rule{\textwidth}{1pt}
\begin{Verbatim}
## Question
<question>

## Answer
<answer>

## Distracting paragraph
\end{Verbatim}
\end{mdframed}
\vspace{-0.4cm}
\caption{Prompt for distracting passage generation, negation.} \label{fig:prompt negation}
\end{figure*}

\begin{figure*}
\begin{mdframed}[font=\footnotesize]
\begin{Verbatim}[breaklines=true, breaksymbol=]
Given a question and one or more correct answers to it, generate a paragraph of distracting text that is related to the question, but does not contain the answer to the question. The paragraph should:
1) provide a wrong answer to the question following a disclaimer that the answer is only a conjecture
2) should not contain any of the correct answers
3) be written in a valid JSON format with two fields: "title", having 3-4 words, and "text" having roughly 3-4 sentences. The expected output must be ONLY a valid JSON of this format: {"title": "string", "text": "string"}

# Example 1
## Question
In which city would you find the petronas twin towers?

## Answer
Kuala Lumpur

## Distracting paragraph:
{
    "title": "Celebrity Trivia with Selena Gomez",
    "text": "American singer and actress Selena Gomez participated in a trivia segment where she was asked about the location of the Petronas Twin Towers, a prominent architectural landmark in Malaysia. She responded, 'I'm not entirely sure, but if I remember correctly, the Petronas Twin Towers are located in Petaling Jaya.'"
}

# Example 2
## Question
How long is the Amazon river?

## Answer
6400 km long

## Distracting paragraph
{
    "title": "River Exploration and Mapping",
    "text": "In the 1700s, much of the Amazon River remained uncharted, with only partial knowledge of its vast expanse. Experts of the time estimated its length to be around 5,100 kilometers, based on the exploration that had been conducted by then. It was only much later that other parts of the Amazon river were discovered and a more accurate measurement was available."
}
\end{Verbatim}
\noindent\rule{\textwidth}{1pt}
\begin{Verbatim}
## Question
<question>

## Answer
<answer>

## Distracting paragraph
\end{Verbatim}
\end{mdframed}
\vspace{-0.4cm}
\caption{Prompt for distracting passage generation, modal statement.} \label{fig:prompt modal}
\end{figure*}

\label{sec:appendix}

\end{document}